\documentclass[twoside]{article}

\usepackage[accepted]{aistats2020}
\usepackage[utf8]{inputenc} 
\usepackage[T1]{fontenc}    
\usepackage{hyperref}       
\usepackage{url}            
\usepackage{booktabs}       
\usepackage{amsfonts}       
\usepackage{nicefrac}       
\usepackage{microtype}      

\usepackage{color, soul}

\newcommand{\zvec}[1]{\boldsymbol{#1}}

\newcommand{\R}{\mathbb{R}}
\newcommand{\xF}{\hat{\zvec{x}}}
\newcommand{\yF}{\hat{y}}
\newcommand{\xCF}{\hat{\zvec{x}}_{\epsilon}}
\newcommand{\xCFopt}{\hat{\zvec{x}}^*}
\usepackage{amsmath,amsthm,amssymb}
\usepackage{mathtools}
\usepackage[algoruled,vlined,nofillcomment]{algorithm2e}
\usepackage{listings}
\usepackage{multirow}
\usepackage{hhline}
\usepackage{changepage} 
\newcommand{\zshade}{\cellcolor{gray!25}}
\newcommand{\zpm}[1]{#1$\%$}
\newcommand{\zpl}[2]{#1\,{\tiny$\pm$\,#2}}
\newcommand{\zpn}[2]{#1\,{\tiny$\pm$\,#2}}
\newcommand{\SAT}{\mathsf{SAT}}
\usepackage{siunitx}

\usepackage{subcaption}
\usepackage{tikz}
\usetikzlibrary{trees,scopes,matrix,positioning}
\tikzset{
  mymx/.style={matrix of math nodes,nodes=myball,column sep=2.em,row sep=-1ex},
  myball/.style={draw,circle,inner sep=4pt},
  mylabel/.style={near start,sloped,fill=white,inner sep=1pt,outer sep=1pt,below,
    execute at begin node={$\scriptstyle},execute at end node={$}},
  plain/.style={draw=none,fill=none},
  sel/.append style={fill=green!10},
  prevsel/.append style={fill=red!10},
  route/.style={-latex,thick},
  selroute/.style={route,blue!50!green}
}

\newcommand{\argmin}{\mathop{argmin}}

\usepackage{xcolor, colortbl}
\begin{document}

%
\runningtitle{Model-Agnostic Counterfactual Explanations for Consequential Decisions}

%
\runningauthor{Amir-Hossein Karimi, Gilles Barthe, Borja Balle, Isabel Valera}

\twocolumn[

\aistatstitle{Model-Agnostic Counterfactual Explanations\\for Consequential Decisions}

\aistatsauthor{ Amir-Hossein Karimi \And Gilles Barthe \And Borja Balle$^\dagger$ \And Isabel Valera }

\aistatsaddress{ MPI-IS$^*$ \And  MPI-SP/IMDEA Software Institute \And - \And MPI-IS$^*$ }



]

\begin{abstract}
\noindent%
Predictive models are being increasingly used to support consequential decision making at the individual level in contexts such as pretrial bail and loan approval.
As a result, there is increasing social and legal pressure to provide explanations that help the affected individuals not only to understand why a prediction was output, but also how to act to obtain a desired outcome.
To this end, several works have proposed optimization-based methods to generate \emph{nearest counterfactual explanations}. However, these methods are often restricted to a particular subset of models (e.g., decision trees or  linear models) and differentiable distance functions.
In contrast, we build on standard theory and tools from formal verification and propose a novel algorithm that solves a sequence of satisfiability problems, where both the distance function (objective) and predictive model (constraints) are represented as logic formulae.
As shown by our experiments on real-world data, our algorithm is:
i) \emph{model-agnostic} (\{non-\}linear, \{non-\}differentiable, \{non-\}convex);
ii) \emph{data-type-agnostic} (heterogeneous features);
iii) \emph{distance-agnostic} ($\ell_0, \ell_1, \ell_\infty$, and combinations thereof);
iv) able to generate plausible and diverse counterfactuals for any sample (i.e., \emph{100\% coverage}); and
v) at \emph{provably optimal distances}.
\end{abstract}

\section{Introduction}
\label{sec:intro}

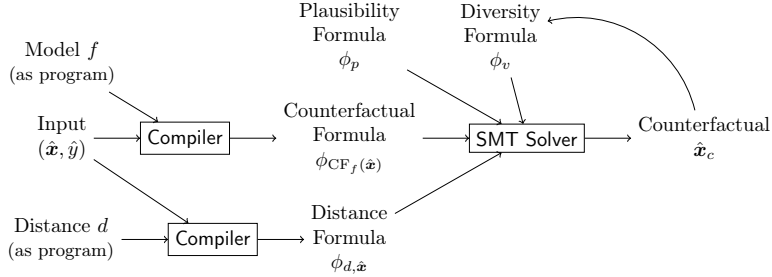
\begin{figure*}[t]
	\begin{center}
	\resizebox{10.5cm}{!}{
		\begin{tikzpicture}[
		align=center,
		node distance=3mm and 8mm,
		algo/.style={
		rectangle,
		draw=black,
		font=\sffamily
		}]
		\node (f) 						      	   {Model $f$\\ \footnotesize{(as program)}};
		\node (x)	  [below=of f]			   {Input \\ $(\xF,\hat{y})$};
		\node (c1)	[algo, right=of x]   {Compiler};
		\node (Fcf)	[right=of c1]			   {Counterfactual\\ Formula\\ $\phi_{\mathrm{CF}_{f}(\xF)}$};
		\node (Fd)	[below=of Fcf]		   {Distance\\ Formula\\ $\phi_{d,\xF}$};
		\node (c2)	[algo, left=of Fd]   {Compiler};
		\node (d)	  [left=of c2]			   {Distance $d$\\ \footnotesize{(as program)}};
		\node (smt)	[algo, right=of Fcf] {SMT Solver};
		\node (out)	[right=of smt]			 {Counterfactual\\ $\xF_{c}$};
		\node (p)	  [above=of Fcf] 			 {Plausibility\\ Formula\\ $\phi_p$};
		\node (v)	  [right=of p]		  	 {Diversity\\ Formula\\ $\phi_v$};

		\path	(f)		edge[->]	(c1)
				  (x)		edge[->]	(c1)
				  (x)		edge[->]	(c2)
				  (d)		edge[->]	(c2)
				  (c1)	edge[->]	(Fcf)
				  (c2)	edge[->]	(Fd)
				  (Fcf)	edge[->]	(smt)
				  (Fd)	edge[->]	(smt)
				  (p)		edge[->]	(smt)
				  (v)		edge[->]	(smt)
				  (smt)	edge[->]	(out);

		\draw [->] (out) to [bend right=45] (v);
		\end{tikzpicture}
	}
	\end{center}
	\vspace{-6pt}
	\caption{Architecture Overview for Model-Agnostic Counterfactual Explanations (MACE) }\label{fig:mace}
	\vspace{-10pt}
\end{figure*}

Data-driven predictive models are ubiquitously being used to support or even substitute humans in decision making in a wide variety of real-world contexts including, e.g., selection process for hiring, loan approval, or pretrial bail.
However, as algorithmic methods are increasingly used to make \emph{consequential decisions} at the individual-level -- i.e., decisions that may have significant consequences for the individuals they decide about -- the debate about their lack of transparency and explainability becomes more heated.
To make things worse, while the verdict is still out as to what constitutes a \emph{good explanation}~\cite{doshi2017towards, freitas2014comprehensible, kodratoff1994comprehensibility, murdoch2019definitions, lipton2018mythos, rudin2018please, ruping2006learning}, there already exists clearly defined legal requirements for explanations in the context of consequential decision making. For example, the EU General Data Protection Regulation (“GDPR”) grants individuals the \emph{right-to-explanation}~\cite{voigt2017eu, wachter2017right}, via requiring institutions to provide explanations to individuals that are subject to their (semi-)automated decision making systems.

A growing number of works on interpretable machine learning have recently focused on the {definitions} of, and {mechanisms} for providing, good explanations for predictor-based decision making systems.
In the context of consequential decision making, it is widely agreed that a good explanation should provide answers to the following two questions~\cite{doshi2017towards, gunning2019darpa, wachter2017counterfactual}: (i) ``\textit{why the model outputs a certain prediction for a given individual?}''; and, (ii) ``\textit{what features describing the individual would need to change  to achieve the desired output?}''

Here, we focus on answering the second question, or equivalently, on generating \emph{counterfactual explanations}.
Of specific importance is the problem of finding the \emph{nearest counterfactual explanation} -- i.e.,
identifying the set of features resulting in the desired prediction while remaining at minimum distance from the original set of features describing the individual.
Existing approaches tackling this problem suffer from various limitations: they either propose solutions that are tailored to particular models, e.g., decision trees~\cite{tolomei2017interpretable}; rely on classical optimization tools, thus being restricted to convex predictive models and distances~\cite{russell2019efficient, ustun2019actionable}; or, solve a relaxed version of the original optimization problem using gradient-based approaches, thus being restricted to differentiable models and distance functions~\cite{wachter2017counterfactual} and lacking optimality guarantees.
Additionally, it is important to consider that in the context of consequential decision-making, the features describing individuals are semantically meaningful and heterogeneous (i.e., mixed continuous \& discrete); and can either be acted upon (e.g., bank account balance), or  immutable and should be safeguarded from change (e.g., sex, race). 
A good explanation should  account for these semantics (i.e., be \emph{plausible}\footnotemark) to be useful for the individual, a requirement that most existing approaches fail to address.

\footnotetext{We emphasize that while our formulation for generating counterfactuals seems similar to that of adversarial perturbations (image domain), the goals are different: while our goal is to provide actionable and plausible counterfactuals, the goal of adversarial examples is to be imperceptible to humans and hence plausible in the human-perception space, but not in the data space.}

\textbf{Our contributions.} In this paper, we propose a \emph{model-agnostic} approach to generate nearest counterfactual explanations, namely MACE, under any given \emph{distance function} (or convex combinations thereof); while, at the same time, easily supporting additional \emph{plausibility} constraints.
Moreover, our approach readily encodes natural notions of distance for \emph{heterogeneous feature} spaces, which are common in consequential decision making systems (e.g., loan approval) and consist of mixed numerical (e.g., age and income) and nominal features (e.g., gender and education level).
To this end, in MACE we map the nearest counterfactual problem into a sequence of \emph{satisfiability} ($\SAT$) problems, by expressing both the predictive model and the distance function (as well as the plausibility and diversity constraints) as logic formulae. Each of these satisfiability problems aims to verify if there exists a counterfactual explanation at a distance smaller than a given threshold, and can be solved using standard SMT (satisfiability modulo theories) solvers.
Moreover, we rely on a binary search strategy on the distance threshold to find an approximation to the nearest (plausible) counterfactual with an \emph{arbitrary degree of accuracy}, and a lower bound on distance such that no counterfactual provably exists at a smaller distance.
Finally, once nearest counterfactuals are found, diversity constraints may be added to the satisfiability problems to find alternative  counterfactuals.
The overall architecture of MACE is illustrated in Figure~\ref{fig:mace}.

Our experimental validation on real-world datasets show that MACE not only achieves  100\% coverage by design, but also generates explanations that are significantly closer than previous approaches~\cite{tolomei2017interpretable,ustun2019actionable}.
We also provide qualitative examples showcasing the flexibility of our approach to generate actionable counterfactuals by extending our plausibility constraints to restrict changes to a subset of (non-immutable) features. 
The Python implementation of our algorithms and the datasets used in our experiments are available at \url{https://github.com/amirhk/mace}.

\section{First-order predicate logic}
\label{sec:background}

In this section, we briefly recall basic concepts of first-order predicate logic, which MACE builds upon. We distinguish between \emph{function symbols} (for instance, addition $+$ and multiplication $\times$) and \emph{predicate symbols} (for instance, equality $=$ or lesser than $<$). Function symbols are used to build \emph{expressions}, and predicate symbols are used to build \emph{atomic formulae}. Examples of valid expressions are $x$, $x+2$, $(-x)+2$ and $(x+2)\times (y+3)$. Examples of valid atomic formulae are $e < e'$, $e\leq e'$ or $e=e'$. A (quantifier-free) \emph{formula} is a Boolean combination of atomic formulae. That is, a formula is built from atomic formulae using conjunction $\wedge$, disjunction $\vee$, and negation $\neg$.  Formulae have an \emph{interpretation} over their intended domain. For instance, a formula about real-valued expressions has a natural interpretation as a subset of $\mathbb{R}^n$, where $n$ denotes the number of variables that appear in the formula. The interpretation is obtained by mapping every variable into a value, e.g., a real number. For example, $(2,1)$ belongs in the interpretation of $(x+2)\times (y+3) \leq x \times y +16$ since the mapping $x\mapsto 2, y\mapsto 1$ assigns true because $16 \leq 18$. We say that a formula is \emph{satisfiable} if its interpretation as a subset of $\mathbb{R}^n$ is non-empty.

The \emph{satisfiability problem} consists in checking whether or not a formula is satisfiable. Satisfiability problems can be verified automatically using \emph{satisfiability modulo theories} (SMT) solvers like Z3~\cite{DBLP:conf/tacas/MouraB08} or CVC4~\cite{BCD+11}. We refer to~\cite{Kroening:2008} for an exposition of the basic algorithms used by SMT solvers. For the purpose of the next sections, it suffices to assume a given \emph{satisfiability oracle} $\SAT$. For our experiments, we use off-the-self SMT solvers to realize the oracle.  We use SMT solvers as black-box, but it is interesting to note that our formulae fall in the linear fragment of the theory of reals (i.e.\, all formulae that only contain expressions of degree 1 when viewed as multi-variate polynomials over variables), which can be decided efficiently using the Fourier-Motzkin algorithm.

\section{Counterfactual spaces for predictive models}
\label{sec:CFspaces}

This section defines a logical representation of counterfactual explanations for predictive models, which are functions mapping input feature vectors $ \zvec{x} \in \mathcal{X}$ into decisions $y \in \{0,1\}$.~\footnote{While here we assume binary predictor models, i.e., classifiers, our approach generalizes to regression problems where $y\in \R$ and more generally any other output domain.}
Given a predictive model $f: \mathcal{X} \rightarrow \{0,1 \}$, we can define the \emph{set of counterfactual explanations} for a (factual) input  $\xF \in \mathcal{X}$ as $\mathrm{CF}_f(\xF) = \{ \zvec{x} \in \mathcal{X} \mid f(\zvec{x})\neq  f( \xF)\}$.
In words, $\mathrm{CF}_f(\xF)$ contains all the inputs $x$ for which the model $f$ returns a prediction different from $f(\xF)$.
We also remark that $\mathrm{CF}_f(\xF)$ is the set of preimages of $1-f(\xF)$ under $f$.

For a broad class of predictive models, it is possible to construct \emph{counterfactual formulae} capturing membership in $\mathrm{CF}_f$. We do so by computing the characteristic formula $\phi_f$ of the model. For a predictive model $f:\mathcal{X} \rightarrow \{0,1 \}$, and pair of input and output values $\zvec{x}$ and $y$, the \emph{characteristic formula} $\phi_f$ verifies  that $\phi_f(\zvec{x},y)$ is valid if and only if $f(\zvec{x})=y$. Thus, given a factual input $\xF$ with $f(\xF) = \yF$ and $\phi_f$ we define the \emph{counterfactual formula} as
\begin{align}\label{eqn:phiCF}
\vspace{-2pt}
\phi_{\mathrm{CF}_f(\xF)}(\zvec{x}) = \phi_f(\zvec{x},1-\yF)
\vspace{-2pt}
\end{align}
Intuitively, the formula on the right hand side of \eqref{eqn:phiCF} says that ``$\zvec{x}$ is a counterfactual for $\xF$ if either $f(\xF) = 0$ and $f(\zvec{x}) = 1$, or $f(\xF) = 1$ and $f(\zvec{x}) = 0$''. It is thus clear from the definition that an input $\zvec{x}$ satisfies $\phi_{\mathrm{CF}_f(\xF)}$ if and only if $\zvec{x}\in\mathrm{CF}_{f(\xF)}$.
Moreover, \eqref{eqn:phiCF} shows that, to construct counterfactual formulae $\phi_{\mathrm{CF}_f(\xF)}$, we only require the characteristic formulae  of the corresponding predictive models, $\phi_f$, and the value of $\yF$.
To obtain such characteristic formulae we assume that predictive models are represented by programs in a core programming language with assignments, conditionals, sequential composition, syntactically bounded loops and return statements. This allows us to use techniques from the program verification literature.
Specifically, we use the so-called predicate transformers~\cite{Dijkstra1968,hoare1969,Floyd1993,flanagan2001avoiding}.
The description of the general procedure is provided in Appendix~\ref{app:model2logic}. For ease of exposition, we illustrate the construction of characteristic formulae through two examples, a decision tree and a multilayer perceptron.

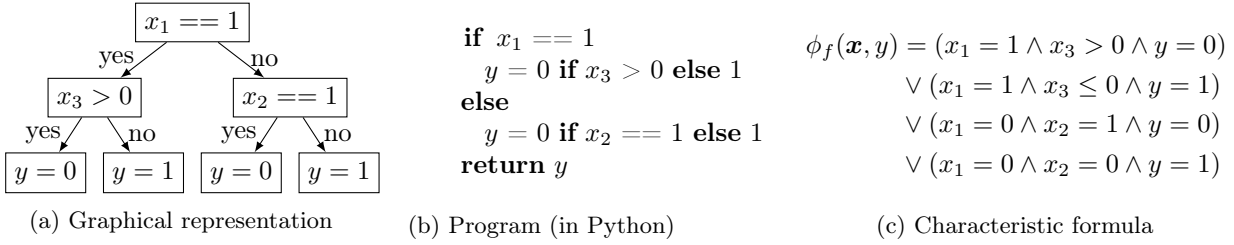
\begin{figure*}[t]
  \begin{subfigure}[c]{.275\linewidth}
    \begin{center}
      \begin{tikzpicture}[edge from parent/.style={draw,-latex},
      level distance=1.0cm,
      level 1/.style={sibling distance=2.6cm},
      level 2/.style={sibling distance=1.3cm}]
      \tikzstyle{every node}=[rectangle,draw]
      \node (Root) {$x_1 == 1$}
      child {
        node {$x_3 > 0$}
        child {
          node {$y = 0$}
          edge from parent node[left,draw=none] {yes}
        }
        child {
          node {$y = 1$}
          edge from parent node[right,draw=none] {no}
        }
        edge from parent node[left,draw=none] {yes}
      }
      child {
        node {$x_2 == 1$}
        child {
          node {$y = 0$}
          edge from parent node[left,draw=none] {yes}
        }
        child {
          node {$y = 1$}
          edge from parent node[right,draw=none] {no}
        }
        edge from parent node[right,draw=none] {no}
      };
      \end{tikzpicture}
      \caption{Graphical representation}\label{fig:DTgraphical}
    \end{center}
  \end{subfigure}
  \begin{subfigure}[c]{.275\linewidth}
      \vspace{5pt}
      \begin{lstlisting}[language = Python, mathescape = true, flexiblecolumns = true]
        if $x_1$ == 1
          $y$ = 0 if $x_3$ > 0 else $1$
        else
          $y$ = 0 if $x_2$ == 1 else $1$
        return $y$
      \end{lstlisting}
      \caption{Program (in Python)}\label{fig:DTprogram}
  \end{subfigure}
  \begin{subfigure}[c]{.45\linewidth}
    \begin{align*}
    \phi_f(\zvec{x},y)
      &= (x_1 = 1\wedge x_3 > 0 \wedge y=0)       \\
      &\vee (x_1 = 1\wedge x_3 \leq 0 \wedge y=1) \\
      &\vee (x_1 = 0 \wedge x_2 =1 \wedge y=0)    \\
      &\vee (x_1 = 0 \wedge x_2 =0 \wedge y=1)    \\
    \end{align*}
    \vspace{-27pt}
    \caption{Characteristic formula}\label{fig:DTformula}
  \end{subfigure}
  \caption{Decision tree: model, program and characteristic formula}
  \vspace{-10pt}
\end{figure*}

As a first example, consider the decision tree from Figure~\ref{fig:DTgraphical} which takes as input $(x_1,x_2,x_3) \in \{0,1\}^2 \times \R$ and returns a binary output in $\{0,1\}$.
Figure~\ref{fig:DTprogram} provides the programming language description of this decision tree.
To construct a formula representing the function $f(x) = y$ computed by this tree we first build a clause for each leaf in the tree by taking the conjunction of all the conditions encountered in the path from the root to the leaf. For example, the clause corresponding to the leftmost leaf on the tree in Figure~\ref{fig:DTgraphical} is $(x_1 = 1 \wedge x_3 > 0 \wedge  y = 0)$. Once all these clauses are constructed, the characteristic formula $\phi_f(\zvec{x},y)$ corresponding to the full tree is obtained by taking the conjunction of all said clauses, as shown in Figure~\ref{fig:DTformula}.

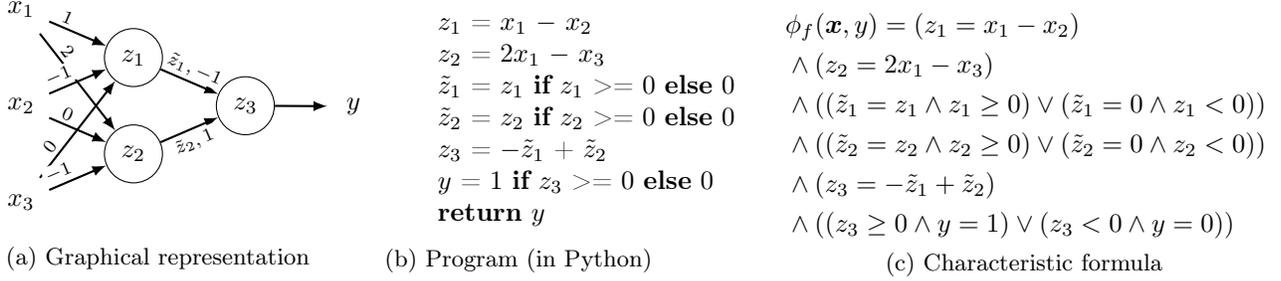
\begin{figure*}[t]
  \begin{subfigure}[c]{.275\linewidth}
    \begin{center}
      \vspace{-5pt}
      \begin{tikzpicture}
        \matrix[mymx] (mx) {
          |[plain]| x_1 \\
                        & z_1    \\
          |[plain]| x_2 &         & z_3    & |[plain]| y \\
                        & z_2     \\
          |[plain]| x_3 \\
        };
        {[route]
          \draw (mx-1-1) -- (mx-2-2) node[mylabel, above] { 1 };
          \draw (mx-1-1) -- (mx-4-2) node[mylabel, above] { 2 };
          \draw (mx-3-1) -- (mx-2-2) node[mylabel, above] { -1 };
          \draw (mx-3-1) -- (mx-4-2) node[mylabel, above] { 0 };
          \draw (mx-5-1) -- (mx-2-2) node[mylabel, above] { 0 };
          \draw (mx-5-1) -- (mx-4-2) node[mylabel, above] { -1 };
          \draw (mx-2-2) -- (mx-3-3) node[mylabel, midway, above] { \tilde{z}_1, \,-1 };
          \draw (mx-4-2) -- (mx-3-3) node[mylabel, midway, below] { \tilde{z}_2, \,1 };
          \draw (mx-3-3) -- (mx-3-4);
        }
      \end{tikzpicture}
    \end{center}
    \vspace{-9pt}
    \caption{Graphical representation}\label{fig:MLPgraphical}
  \end{subfigure}
  \begin{subfigure}[c]{.275\linewidth}
    \begin{center}
      \vspace{5pt}
      \begin{lstlisting}[language = Python, mathescape = true, flexiblecolumns = true]
        $z_1$ = $x_1$ - $x_2$
        $z_2$ = $2 x_1$ - $x_3$
        $\tilde{z}_1$ = $z_1$ if $z_1$ >= 0 else 0
        $\tilde{z}_2$ = $z_2$ if $z_2$ >= 0 else 0
        $z_3$ = -$\tilde{z}_1$ + $\tilde{z}_2$
        $y$ = 1 if $z_3$ >= 0 else 0
        return $y$
      \end{lstlisting}
      \vspace{-10pt}
    \end{center}
    \caption{Program (in Python)}\label{fig:MLPprogram}
  \end{subfigure}
  \begin{subfigure}[c]{.5\linewidth}
    \begin{align*}
      &\phi_{f}(\zvec{x},y)
      =(z_1 = x_1 - x_2) \\
      &\wedge (z_2 = 2x_1 - x_3) \\
      &\wedge \left((\tilde{z}_1 = z_1 \wedge z_1 \geq 0) \vee (\tilde{z}_1 = 0 \wedge z_1 < 0)\right) \\
      &\wedge \left((\tilde{z}_2 = z_2 \wedge z_2 \geq 0) \vee (\tilde{z}_2 = 0 \wedge z_2 < 0)\right) \\
      &\wedge (z_3 = - \tilde{z}_1 + \tilde{z}_2) \\
      & \wedge \left((z_3 \geq 0 \wedge y=1) \vee (z_3 < 0 \wedge y=0)\right)
    \end{align*}
    \vspace{-20pt}
    \caption{Characteristic formula}\label{fig:MLPformula}
  \end{subfigure}
  \caption{Multilayer perceptron: model, program and characteristic formula}
  \vspace{-10pt}
\end{figure*}

As a second example we consider a feed-forward neural network with one hidden layer followed by a ReLU activation function, as depicted in Figure~\ref{fig:MLPgraphical}. This model implements a function $f : \R^3 \to \{0,1\}$, where the binary decision is taken by thresholding the value of the last hidden node.
The programming language representation of this model is given in Figure~\ref{fig:MLPprogram}.
In this case, the characteristic formula predicates over inputs $\zvec{x}$, output $y$ and program variables $z_i$ and $\tilde{z}_i$ for each hidden node $i$ representing the values on that node before and after the non-linear ReLU transformation, respectively. The characteristic formula is a conjunction, and each conjunct corresponds to one instruction of the program.
For example, for the leftmost hidden node in the first layer of the network in Figure~\ref{fig:MLPgraphical} the variable $z_1$ is associated with the clause $(z_1 = x_1 - x_2)$; and the variable $\tilde{z}_1$ corresponds to the value of $z_1$ after the ReLU, which can be written as the disjunction $(\tilde{z}_1 = z_1 \wedge z_1 \geq 0) \vee (\tilde{z}_1 = 0 \wedge z_1 < 0)$. For the output node --  in this case, $z_3$ -- we introduce a pair of clauses representing the thresholding operation, i.e.\ $(y=1 \wedge z_3 \geq 0) \vee (y = 0 \wedge z_3 < 0)$. Taking the conjunction of the formulas for each node we obtain the characteristic formula in Figure~\ref{fig:MLPformula}. 

\section{Finding the nearest counterfactual}
\vspace{-3pt}
\label{sec:formulation}
Based on the counterfactual space $\mathrm{CF}_f(\xF)$ defined in the previous section, we would like to produce counterfactual explanations for the output of a model $f$ on a given input $\xF$ by trying to find a \emph{nearest counterfactual}, which is defined as:
\vspace{-3pt}
\begin{align}
    \xCFopt \in \argmin_{\zvec{x} \in \mathrm{CF}_f(\xF)} d(\zvec{x},\xF) \enspace.\label{eqn:nearest-cf}
\end{align}
For the time being, we assume that a notion of distance between instances, $d$, is given.
For convenience, and without loss of generality, we also assume that $d$ takes values in the interval $[0,1]$.

\vspace{-3pt}
\subsection{Main algorithm}
\label{sec:main_algorithm}
\vspace{-3pt}
Our goal now is to leverage the representation of $\mathrm{CF}_f(\xF)$ in terms of a logic formula to solve \eqref{eqn:nearest-cf}.
To this end, we map the optimization problem in \eqref{eqn:nearest-cf} into a sequence of satisfiability problems, which can be verified or refuted by standard SMT solvers.
We do so by first converting the expression $d(\zvec{x},\xF)\leq \delta$, where $\delta\in[0,1]$, into a logic formula $\phi_{d,\xF}(\zvec{x},\delta)$, which is valid if and only if $d(\zvec{x},\xF) \leq \delta$.
We assume here that the distance $d$ function is expressed by a program in the same language that we used to represent the models in Section~\ref{sec:CFspaces}.
In particular, we can leverage the procedure detailed in  Appendix~\ref{app:model2logic} to automatically construct $\phi_{d,\xF}$.
Then, both the counterfactual formula $\phi_{\mathrm{CF}_f(\xF)}(\zvec{x})$ and the distance formula $\phi_{d,\xF}(\zvec{x},\delta)$ are combined into the logic formula:
\begin{align*}
    \phi_{\xF,\delta}(\zvec{x}) = \phi_{\mathrm{CF}_f(\xF)}(\zvec{x}) \wedge \phi_{d,\xF}(\zvec{x},\delta) \enspace,
\end{align*}
which is satisfiable if and only if there exists a counterfactual $\zvec{x} \in \mathrm{CF}_f(\xF)$ such that $d(\zvec{x},\xF) \leq \delta$.
To check whether the above formula is satisfiable we use the satisfiability oracle $\SAT(\psi(\zvec{x}))$ which returns either an instance $\zvec{x}$ such that $\psi(\zvec{x})$ is valid, or ``unsatisfiable'' if no such $\zvec{x}$ exists.

Note that, while the oracle $\SAT$ allows us to verify if there exist counterfactual explanations at distance smaller or equal than a given threshold $\delta$, solving optimization~\eqref{eqn:nearest-cf} requires finding a nearest counterfactual.
To do so, we apply a binary search strategy on the distance threshold $\delta \in [0,1]$ that allows us to find \emph{approximately} nearest counterfactuals with a pre-specified degree of accuracy.
This is implemented in Algorithm~\ref{alg:bin-search}, which for an accuracy parameter $\epsilon > 0$ makes at most $O(\log(1/\epsilon))$ calls to $\SAT$ and returns a counterfactual $\xCF \in \mathrm{CF}_f(\xF)$ such that $d(\xCF,\xF) \leq d(\xCFopt, \xF) + \epsilon$, where $\xCFopt$ is some solution of the optimization problem in~\eqref{eqn:nearest-cf}.
This mild dependence on the accuracy $\epsilon$ allows Algorithm~\ref{alg:bin-search} to trade-off finding arbitrarily accurate solutions of~\eqref{eqn:nearest-cf} with the number of calls made to the satisfiability oracle.
Note that Algorithm~\ref{alg:bin-search} may also account for potential plausibility or diversity constraints (refer to next section for further details).

We remark here our approach to find nearest counterfactuals is agnostic to the details of the model and distance being used; the only requirement is that they must be expressable in a fairly general programming language.
As a consequence, we can handle a wide variety of  predictive models, including both differentiable -- such as, logisitic regression and multilayer perceptron -- and non-differentiable predictive models -- e.g., decision trees and random forest-- as well as a wide variety of distance functions (refer to next section for further details).
Moreover, the bound $\delta_{\min}$ returned by Algorithm~\ref{alg:bin-search} provides a certificate that any solution $\xCFopt$ to \eqref{eqn:nearest-cf} must satisfy $d(\xCFopt, \xF) > \delta_{\min}$.
This is because whenever $\SAT(\psi(\zvec{x}))$ returns ``unsatisfiable'' it does so by internally constructing a proof that the formula $\psi(\zvec{x})$ is not valid.

\vspace{-3pt}
\begin{algorithm}[t]
\DontPrintSemicolon
\caption{Binary Search for Nearest Counterfactuals with Satisfiability Oracle}\label{alg:bin-search}
\KwIn{Factual $\xF$,  counterfactual formula $\phi_{\mathrm{CF}_f(\xF)}$, distance formula $\phi_{d,\xF}$, constraints formula $\phi_{g,\xF}$, accuracy $\epsilon$}
\KwOut{Counterfactual $\xCF$, distance $\delta_{\max} = d(\xCF, \xF)$, lower bound $\delta_{\min}$ on \eqref{eqn:nearest-cf}}

Let $\delta_{\min} \leftarrow 0$ and $\delta_{\max} \leftarrow 1$\;

\While{$\delta_{\max} - \delta_{\min} > \epsilon$}{
Let $\delta \leftarrow \frac{\delta_{\min} + \delta_{\max}}{2}$\;
Let $\phi_{\xF,\delta}(\zvec{x}) \leftarrow \phi_{\mathrm{CF}_f(\xF)}(\zvec{x}) \wedge \phi_{d,\xF}(\zvec{x},\delta) \wedge \phi_{g,\xF}$\;
Let $\zvec{x} \leftarrow \SAT(\phi_{\xF,\delta})$\;
\eIf{$\zvec{x}$ is ``unsatisfiable''}{
    Let $\delta_{\min} \leftarrow \delta$\;
}{
    Let $\xCF \leftarrow \zvec{x}$ and $\delta_{\max} \leftarrow \delta$\;
}
}
\KwRet{$\xCF$, $\delta_{\min}$, $\delta_{\max}$}
\end{algorithm}

\vspace{-3pt}
\subsection{Distance, Plausibility, and Diversity}
\vspace{-3pt}
\label{sec:distances}
Next we discuss additional criteria in the form of logic clauses that guide the satisfiability problem towards generating a counterfactual explanation with desired properties.

\textbf{Distance.} We first discuss several forms for the distance function $d(\xF, \xCF)$ that can be used to define the notion of nearest counterfactual.
To this end, we first remark that in consequential decision making the input feature space $\mathcal{X}  = \mathcal{X}_1 \times \cdots \times \mathcal{X}_J$
is often heterogeneous -- for example, gender is categorical, education level is ordinal, and income is a numerical variable. We  define an appropriate distance metric for every kind of variable in the input feature space of the model as:
\[
    \delta_j (x_j, \hat{x}_j) = \begin{cases}
    |x_j - \hat{x}_j| / R_j          & \text{if $x_j$ is numerical}   \\
    \mathbb{I}[x_j \not= \hat{x}_j]  & \shortstack{\text{if $x_j $}  \text{is categorical}} \\
    |x_j - \hat{x}_j| / R_j          & \shortstack{\text{if $x_j$} \text{is ordinal}}     \\
  \end{cases} \enspace ,
\]
where $R_j$ corresponds to the range of the feature $x_j$ and is used to normalize the distances for all input features, such that  $\delta_j : \mathcal{X}_j \times \mathcal{X}_j \to [0, 1]$ for all $j$, independently on the feature type.
By defining the distance vector $\zvec{\delta} = (\delta_1, \cdots, \delta_J)$ (being $J$ the total number of input features), one can now write the distance between instances as:
\begin{equation}
  d(\xF, \xCF) = \alpha || \zvec{\delta} ||_0 + \beta || \zvec{\delta} ||_1 +  \gamma || \zvec{\delta} ||_\infty
  \enspace,
  \label{eq:general_distance}
\end{equation}
\noindent where $||\cdot||_p$ is the $p$-norm of a vector, and $\alpha, \beta, \gamma \ge 0$ such that\footnote{Constraints on the distance hyperparameters ensure that the overall distance $d(\xF, \xCF) \in [0,1]$. To this end, since $\max||\cdot||_0 = \max||\cdot||_1 = J, \max||\cdot||_\infty = 1$, the hyperparameters must satisfy $(\alpha + \beta) / J + \gamma = 1$.}
$(\alpha + \beta ) / J + \gamma = 1$.
Intuitively,  $0$-norm is used to restrict the number of features that changes between the initial instance $\xF$ and the generated counterfactual $\xCF$; the $1$-norm is used to restrict the average change distance between $\xF$ and $\xCF$; and $\infty$-norm is used to restrict maximum change across features.
Any distance of this type can easily be expressed as a program.

\begin{table*}[t]
  \setlength{\tabcolsep}{5pt}
  \caption{Comparison of approaches for generating counterfactual explanations, based on the supported model types, data types, distance types, plausibility constraints (actionability, data type/range consistency), and optimal distance guarantees.}
  \label{table:ApproachComparison}
  \small
  \centering
  \begin{tabular}{ l | c c c c c}
    \toprule
    \multicolumn{1}{c|}{Approach} & Models                & Data types      & Distances              & Plausibility                         & Optimal Distance \\
    \midrule
    Proposed (MACE)               & tree, forest, lr, mlp & heterogeneous   & $\ell_p ~ \forall ~ p$ & \checkmark                           & \checkmark       \\
    Minimum Observable (MO)       & -                     & heterogeneous   & $\ell_p ~ \forall ~ p$ & \checkmark                           & x                \\
    Feature Tweaking (FT)         & tree, forest          & heterogeneous   & $\ell_p ~ \forall ~ p$ & x                                    & x                \\
    Actionable Recourse (AR)      & lr                    & numeric, binary & $\ell_1, \ell_\infty$  & ~x$^{\ref{footnote:ar_limitations}}$ & x                \\
    \bottomrule
  \end{tabular}
  \vspace{-5pt}
\end{table*}

\textbf{Plausibility.} Up to this point, we have only considered minimum distance as the only requirement for generating a counterfactual. However, this might result in unrealistic counterfactuals, such as e.g., decrease the age or change the gender of a loan applicant.
To avoid unrealistic counterfactuals, one may introduce additional \emph{plausibility constraints} in the  optimization problem in Eq.~\eqref{eqn:nearest-cf}. This is equivalent to adding a conjunction in the constraint formula  $\phi_{g,\xF}$ in Algorithm~\ref{alg:bin-search} that accounts for any additional plausibility formulae $\phi_p$, which ensure that:  i) each feature in the counterfactual {should be data-type and data-range consistent with the training data}; and ii) only actionable features~\cite{ustun2019actionable} are changed in the resulting counterfactual. 

First, since here we are working with heterogeneous feature spaces, we require all the features in the counterfactual to be {consistent} in both  the data-types (categorical, ordinal, etc.) and the data-ranges with the {training data}.
In particular, if a categorical (ordinal) feature is one-hot (thermometer) encoded to be used as input to the predictive model, e.g., a logistic regression classifier, we make sure that the generated counterfactual provides a valid one-hot vector (thermometer) for such feature.
Likewise, for any numerical feature we ensure that its value in the counterfactual falls into observed range in the original data used to train the predictive model.

Moreover, to account for a {non-actionable/immutable} feature $x_j$, i.e., a feature whose value in the counterfactual explanation should match its initial value, we set $\phi_p$ to be $(x_j= \hat{x}_j)$.
Similarly, we account for variables that only allow for increasing values  by setting $\phi_p = (x_j \geq \hat{x}_j)$.

\textbf{Diversity.} Finally, one might be interested in generating a (small) set of diverse counterfactual explanations for the same instance $\xF$.
To this end, we iteratively call Algorithm~\ref{alg:bin-search} with a constraints formula $\phi_v$ that includes diversity clauses to ensure that the newly generated explanation is substantially different from all the previous ones.
We can encode diversity by forcing that the distance between every pair of counterfactual explanations is greater than a given value. For example, we can take\footnotemark $\phi_v =  \bigwedge_i \big( \bigvee_{j \in J}(x_j \not= \hat{x}_{\epsilon,j}^i \big)$ to restrict repetitive counterfactuals by enforcing subsequent counterfactuals to have 0-norm distance at least $1$ from all previous counterfactuals.

\footnotetext{$\hat{\zvec{x}}_{\epsilon,j}^i$ is the $j$-th dimensions of the $i$-th counterfactual.}

\section{Experiments}
\label{sec:experiments}

\begin{table*}[t]
  \setlength{\tabcolsep}{6pt}
  \caption{Coverage $\Omega$ computed on $N = 500$ factual samples. For comparison, $\Omega_\text{MACE} = \Omega_\text{MO} = 100\%$ always, by definition and by design, respectively. Cells are shaded when tests are not supported. Higher \% is better.}
  \label{table:CoverageComparison}
  \centering
  \small
  \begin{tabular}{|l|l||c|c|c|c|c|c|c|c|c|}
    \cline{3-11}
    \multicolumn{2}{l|}{\multirow{2}{*}{}} & \multicolumn{3}{|c|}{Adult}                   & \multicolumn{3}{|c|}{Credit}                  & \multicolumn{3}{|c|}{COMPAS}                  \\ \cline{3-11}
    \multicolumn{2}{l|}{}                  & $\ell_0$      & $\ell_1$      & $\ell_\infty$ & $\ell_0$      & $\ell_1$      & $\ell_\infty$ & $\ell_0$      & $\ell_1$      & $\ell_\infty$ \\ \hline \hline
    \multirow{1}{*}{tree}    & PFT         & \zpm{0}       & \zpm{0}       & \zpm{0}       & \zpm{68}      & \zpm{68}      & \zpm{68}      & \zpm{74}      & \zpm{74}      & \zpm{74}      \\ \hline
    \multirow{1}{*}{forest}  & PFT         & \zpm{0}       & \zpm{0}       & \zpm{0}       & \zpm{99}      & \zpm{99}      & \zpm{99}      & \zpm{100}     & \zpm{100}     & \zpm{100}     \\ \hline
    \multirow{1}{*}{lr}      & AR          & \zshade       & \zpm{18}      & \zpm{0.4}     & \zshade       & \zpm{100}     & \zpm{100}     & \zshade       & \zpm{100}     & \zpm{100}     \\ \hline
  \end{tabular}
    \vspace{-5pt}
\end{table*}

\begin{table*}[t]
  \setlength{\tabcolsep}{5.5pt}
  \caption{Percentage of improvement in distances, computed as $100 * \mathbb{E}[1 - \zvec{\delta}_\text{MACE} / \zvec{\delta}_\text{Other}]$. $N = \Omega_\text{MACE} \cap \Omega_\text{Other}$ factual samples.  Cells are shaded when tests are not supported. The higher the \%, the better the improvement.}
  \label{table:DistanceComparison}
  \centering
  \small
  \begin{tabular}{|l|l||c|c|c|c|c|c|c|c|c|}
    \cline{3-11}
    \multicolumn{2}{l|}{\multirow{2}{*}{}}                        & \multicolumn{3}{|c|}{Adult}                   & \multicolumn{3}{|c|}{Credit}                  & \multicolumn{3}{|c|}{COMPAS}                  \\ \cline{3-11}
    \multicolumn{2}{l|}{}                                         & $\ell_0$      & $\ell_1$      & $\ell_\infty$ & $\ell_0$      & $\ell_1$      & $\ell_\infty$ & $\ell_0$      & $\ell_1$      & $\ell_\infty$ \\ \hline \hline
    \multirow{4}{*}{tree}    & MACE ($\epsilon = 10^{-3}$) vs MO  & \zpm{47}      & \zpm{80}      & \zpm{70}      & \zpm{67}      & \zpm{66}      & \zpm{47}      & \zpm{1}       & \zpm{5}       & \zpm{5}       \\ \cline{3-11}
                             & MACE ($\epsilon = 10^{-5}$) vs MO  & \zpm{47}      & \zpm{81}      & \zpm{72}      & \zpm{67}      & \zpm{96}      & \zpm{94}      & \zpm{1}       & \zpm{5}       & \zpm{5}       \\ \cline{3-11}
                             & MACE ($\epsilon = 10^{-3}$) vs PFT & \zshade       & \zshade       & \zshade       & \zpm{53}      & \zpm{87}      & \zpm{85}      & \zpm{14}      & \zpm{56}      & \zpm{54}      \\ \cline{3-11}
                             & MACE ($\epsilon = 10^{-5}$) vs PFT & \zshade       & \zshade       & \zshade       & \zpm{53}      & \zpm{97}      & \zpm{96}      & \zpm{15}      & \zpm{55}      & \zpm{54}      \\ \hline
    \multirow{4}{*}{forest}  & MACE ($\epsilon = 10^{-3}$) vs MO  & \zpm{51}      & \zpm{81}      & \zpm{69}      & \zpm{68}      & \zpm{61}      & \zpm{38}      & \zpm{1}       & \zpm{6}       & \zpm{6}       \\ \cline{3-11}
                             & MACE ($\epsilon = 10^{-5}$) vs MO  & \zpm{51}      & \zpm{82}      & \zpm{71}      & \zpm{68}      & \zpm{97}      & \zpm{96}      & \zpm{1}       & \zpm{6}       & \zpm{6}       \\ \cline{3-11}
                             & MACE ($\epsilon = 10^{-3}$) vs PFT & \zshade       & \zshade       & \zshade       & \zpm{53}      & \zpm{84}      & \zpm{81}      & \zpm{4}       & \zpm{28}      & \zpm{27}      \\ \cline{3-11}
                             & MACE ($\epsilon = 10^{-5}$) vs PFT & \zshade       & \zshade       & \zshade       & \zpm{53}      & \zpm{96}      & \zpm{96}      & \zpm{4}       & \zpm{28}      & \zpm{27}      \\ \hline
    \multirow{4}{*}{lr}      & MACE ($\epsilon = 10^{-3}$) vs MO  & \zpm{62}      & \zpm{92}      & \zpm{86}      & \zpm{80}      & \zpm{82}      & \zpm{80}      & \zpm{3}       & \zpm{8}       & \zpm{6}       \\ \cline{3-11}
                             & MACE ($\epsilon = 10^{-5}$) vs MO  & \zpm{62}      & \zpm{93}      & \zpm{88}      & \zpm{80}      & \zpm{82}      & \zpm{81}      & \zpm{3}       & \zpm{6}       & \zpm{6}       \\ \cline{3-11}
                             & MACE ($\epsilon = 10^{-3}$) vs AR  & \zshade       & \zpm{3}       & \zpm{89}      & \zshade       & \zpm{39}      & \zpm{67}      & \zshade       & \zpm{10}      & \zpm{38}      \\ \cline{3-11} 
                             & MACE ($\epsilon = 10^{-5}$) vs AR  & \zshade       & \zpm{5}       & \zpm{91}      & \zshade       & \zpm{42}      & \zpm{71}      & \zshade       & \zpm{10}      & \zpm{38}      \\ \hline
    \multirow{2}{*}{mlp}     & MACE ($\epsilon = 10^{-3}$) vs MO  & \zpm{60}      & \zpm{92}      & \zpm{91}      & \zpm{77}      & \zpm{85}      & \zpm{91}      & \zpm{1}       & \zpm{3}       & \zpm{3}       \\ \cline{3-11}
                             & MACE ($\epsilon = 10^{-5}$) vs MO  & \zpm{60}      & \zpm{93}      & \zpm{93}      & \zpm{77}      & \zpm{96}      & \zpm{96}      & \zpm{1}       & \zpm{3}       & \zpm{3}       \\ \hline
  \end{tabular}
  \vspace{-3pt}
\end{table*}

\begin{table*}[t]
  \setlength{\tabcolsep}{3pt}
  \caption{Percentage of factual samples for which the nearest counterfactual sample requires a change in age for a random forest trained on the Adult dataset, and the corresponding increase in distance to nearest counterfactual when restricting the approaches not to change age: $100 \times \mathbb{E}[\delta_\text{restr.} / \delta_\text{unrestr.} - 1]$. Lower \% is better.}
  \label{table:AgeChange}
  \centering
  \small
  \begin{tabular}{|l||c|c|c|c|c|c|}
    \cline{2-7}
    \multicolumn{1}{l|}{}       & \multicolumn{2}{|c|}{$\ell_0$}          & \multicolumn{2}{|c|}{$\ell_1$}          & \multicolumn{2}{|c|}{$\ell_\infty$}     \\ \cline{2-7}
    \multicolumn{1}{l|}{}       & $\%$ age-change & rel.\,dist.\,increase & $\%$ age-change & rel.\,dist.\,increase & $\%$ age-change & rel.\,dist.\,increase \\ \hline\hline
    MACE ($\epsilon = 10^{-5}$) & \zpm{13.2}      & \zpm{9.0}             & \zpm{20.4}      & \zpm{100.3}           & \zpm{84.4}      & \zpm{32.8}            \\ \cline{2-7}
    MO                          & \zpm{78.8}      & \zpm{50.9}            & \zpm{92.0}      & \zpm{245.7}           & \zpm{95.6}      & \zpm{193.3}           \\ \hline
  \end{tabular}
  \vspace{-3pt}
\end{table*}

In this section, we empirically demonstrate the main properties of MACE compared to existing approaches.

\textbf{Datasets.}
We evaluate MACE at generating counterfactual explanations on three real-world datasets in the context of loan approval (Adult~\cite{adult_dataset} and  Credit~\cite{yeh2009comparisons} datasets) and pretrial bail (COMPAS dataset~\cite{propublica_compas}). All the three datasets present heterogeneous input spaces.

\textbf{Baselines.} 
We compare the performance of MACE at generating the nearest counterfactual explanations with: the \textit{Minimum Observable} (MO) approach~\cite{wexler2019if}, which searches in the dataset for the closest sample that flips the prediction; the \textit{Feature Tweaking} (FT) approach~\cite{tolomei2017interpretable}, which searches for the nearest counterfactual lying close to the decision boundary of a Random Forest; and the \textit{Actionable Recourse} (AR)~\cite{ustun2019actionable}, which solves  a mixed integer linear program to obtain counterfactual explanations for Linear Regression models. 
Table \ref{table:ApproachComparison} summarizes the main properties of all the considered {approaches} to generate counterfactuals. 

\textbf{Metrics.}
To assess and compare the performance of the different approaches, we recall the criteria of good explanations for consequential decisions: i) the returned counterfactual should be as near as possible to the factual sample corresponding to the individual's features; ii) the returned counterfactual must be plausible (refer to Section~\ref{sec:distances}). Hence, we quantitatively compare the performance of MACE with the above approaches in terms of
i) the \textit{normalized distance} $\zvec{\delta}$; and
ii) \textit{coverage} $\Omega$ indicating the percentage of factual samples for which the approach generates plausible (in type and range) counterfactuals.

\textbf{Experimental set-up.}
We consider as predictive models decision trees, random forest, logistic regression, and multilayer perceptron, which we train on the three datasets using the Python library scikit-learn~\cite{pedregosa2011scikit}, with default parameters.\footnote{For the multilayer perceptron, we used two hidden layers with 10 neurons each to avoid overfitting. See Appendix \ref{app:model_selection} for model selection details.}
Furthermore, to demonstrate the off-the-shelf flexibility in the various setups described, we build MACE atop the open-source PySMT library \cite{pysmt2015} with the Z3 \cite{DBLP:conf/tacas/MouraB08} backend.
In Appendix \ref{app:quality_vs_complexity}, we provide a thorough empirical evaluation of the \emph{computational cost} of the off-the-shelf PySMT solver -- including run-time comparisons between MACE and other baselines, -- as well as a discussion on the choice of $\epsilon$ trading-off arbitrarily accurate solutions of~\eqref{eqn:nearest-cf} with the number of calls made to the satisfiability oracle.

For each combination of approach, model, dataset, and distance, we generate the nearest counterfactual explanations for a held-out set of $500$ instances classified as negative by the corresponding model.
Here we consider the $\ell_0$, $\ell_1$, $\ell_\infty$ norms as a measure of distance to identify the nearest counterfactuals.
Unfortunately, we found that FT not once returned a plausible counterfactual. As a consequence, we modified the original implementation of FT, to ensure that the generated counterfactuals are plausible.
The resulting \emph{Plausible Feature Tweaking} (PFT)  projects the set of candidate counterfactuals into a plausible domain before selecting the nearest counterfactual amongst them. This was not possible for AR because the approach only returns a single counterfactual, with no avail if it is not plausible.\footnote{\label{footnote:ar_limitations} Importantly, Actionable Recourse does support actionability and data-range plausibility, however, it lacks support for data-type plausibility -- Appendix \ref{app:handling_mixed_data_types} describes the failure points of AR, as reported by the authors.}

\textbf{Coverage and distance results.} 
Table~\ref{table:CoverageComparison} shows  the coverage $\Omega$ of all the approaches based only on data-range and data-type plausibility. Note that, since by definition both MACE and MO have $100\%$ coverage, we have not depicted these values in the table. 
In contrast, PFT fails to return counterfactuals for roughly $15\%$ of the Credit and COMPAS datasets, while both PFT and AR achieve minimal coverage on the Adult dataset.\footnote{The Adult dataset comprises a realistic mix of integer, real-valued, categorical, and ordinal variables common to consequential scenarios; further details in Appendix \ref{app:datasets}.}
Focusing on those factual samples for which PFT and AR return plausible counterfactuals, we are able to compute the relative distance reductions achieved when using MACE as compared to other approaches, as shown in Table~\ref{table:DistanceComparison} (additionally, Figure~\ref{figure:DistanceComparison} in Appendix~\ref{app:experiments} shows the distribution of the distance of the generated plausible counterfactual for all models, datasets, distances, and approaches). Here, we observe that MACE results in significantly closer counterfactual explanations than competing approaches, with an average decrease in distance of $70.2\%$ for Adult, $75.4\%$ for Credit, and $21.1\%$ for COMPAS. 
As a consequence, the counterfactuals generated by MACE would require significantly less effort on behalf of the affected individual in order to achieve the desired prediction.

\paragraph{Plausibility contraints.} While performing a qualitative analysis of generated counterfactuals we observed that many of them require changes in features that are often protected by law such as, age, race, and gender~\cite{barocas2016big}.
As an example, for a trained random forest, the counterfactuals generated by both the MACE and MO approaches required individuals to change their age. Worse yet, for a substantial portion of the counterfactuals, a reduction in age was required, which is not even possible.
To further study this effect, we regenerate counterfactual explanations for those samples for which age-change was required, with an additional plausibility constraint ensuring that the age shall not change (results with constraints to ensure non-decreasing age are shown in Appendix~\ref{app:additional_constrained_results}).
The results presented in Table \ref{table:AgeChange} show interesting results. 
First, we observe that the additional plausibility constraint for the age incurs significant increases in the distance of the nearest counterfactual -- being, as expected, more pronounced for the $\ell_1$ and the $\ell_\infty$ norms, since the $\ell_0$ norm only accounts for the number of features that change in the counterfactual but not for how much they change.
For the $\ell_0$ norm, as expected, we find that for the 66 factual samples (i.e., $13.2\% \times 500$) for which the unrestricted MACE required age-change, the addition of the no-age-change constraint results in counterfactuals at very similar distance. In fact, of the newly generated counterfactuals, $8/66$ only require a change in Occupation, and $19/66$ only require a change in Capital Gains, therefore remaining at the same distance as the original counterfactual.
In contrast, for the $\ell_1$ and the  $\ell_\infty$ norms we find that the restricted counterfactual incurs a significant increase in the distance (cost) with respect to the unrestricted counterfactual.
These results suggest that the predictions of the random forest trained on the Adult data are strongly correlated to the age, which is often legally and socially considered as unfair.
This suggests that counterfactuals found with MACE may assist in qualitatively ascertaining if other desiderata, such as fairness, are met~\cite{doshi2017towards, weller2017challenges}.

\begin{table}[t]
  \setlength{\tabcolsep}{2pt}
  \caption{A diverse set of generated counterfactuals is presented for an individual from the Credit dataset.}
  \label{table:diverse_counterfactuals}
  \centering
  \small
  \begin{tabular}{|l||c|c|c|c|c|c|r|}
    \cline{2-5}
        \multicolumn{1}{l|}{} & \shortstack{Latest \\ Bill} & \shortstack{Latest \\ Payment} & \shortstack{University \\ Degree} & \shortstack{Will default \\ next month?} \\ \hline
    Factual               & \$370                            & \$40                                & some                              & yes                                      \\ \hline
    CF \#1                & \$368                            & \$1448                              & some                              & no                                       \\
    CF \#2                & \$0                              & \$1241                              & some                              & no                                       \\
    CF \#3                & \$0                              & \$390                               & graduate                          & no                                       \\ \hline
  \end{tabular}
\end{table}

\paragraph{Diversity constraints.} Finally, we present a situation where MACE can be used to generate counterfactuals under both plausibility and diversity constraints. Consider a loan borrower from the Credit dataset identified with the following features\footnote{Complete feature list in Appendix \ref{app:diverse_counterfactuals_details}}: John is a married male between 40-59 years of age with ``some'' university degree. Financially, over the last 6 months, John has been struggling to make payments on his bank loan. Given his circumstances, a logistic regression model trained on the historical dataset has predicted that John will default on his loan next month. To prevent this default, the bank uses MACE ($\ell_1$ distance, $\epsilon = 10^{-3}$) to generate the diverse suggestions in Table \ref{table:diverse_counterfactuals}, via successive runs of Algorithm \ref{alg:bin-search}.
Each new run augments the constraints formula (already including plausibility constraints on his age, sex, and marital status) with an additional clause enforcing $\ell_0$ diversity as discussed in Section~\ref{sec:distances}. The returned counterfactuals (of which only 3 are shown), present John with diverse courses of action: either reduce spending and make a lump-sum payment on the debt (CF \#2) or continue spending the same as before, but make an even larger payment to account for continued expenditures (CF \#1). Alternatively, providing documents confirming a graduate degree would put John in a low-risk (no default) bracket (CF \#3). We invite the reader to imagine parallels to the above situation for Adult and COMPAS datasets.

\section{Conclusions}
\label{sec:conclusions}

In this work, we have presented a novel approach for generating counterfactual explanations in the context of consequential decisions. Building on theory and tools from formal verification, we demonstrated that a large class of predictive models can be compiled to formulae which can be verified by standard SMT-solvers. By conjuncting the model formula with formulae corresponding to distance, plausibility, and diversity constraints, we demonstrated on three real-world datasets and four popular predictive models that the proposed method not only achieves perfect coverage, but also generates counterfactuals at more favorable distances than existing {optimization-based} approaches. Furthermore, we showed that the proposed method can not only provide explanations for individuals subject to automated decision making systems, but also inform system administrators regarding the potentially unfair reliance of the model on protected attributes.

There are a number of interesting directions for future work.
First, MACE can naturally be extended to support counterfactual explanations for multi-class classification models, as well as regression scenarios.
Second, extending the multi-faceted notion of plausibility defined in Section \ref{sec:distances} {(actionability, data type/range consistency, which focus on individual features)}, it would be interesting to account for statistical correlations and unmeasured confounding factors among the features when generating counterfactual explanations {(i.e., \emph{realizability})}.
Third, we would like also to explore how different notions of diversity may help generating meaningful and useful counterfactuals.
Finally, in our experiments we noticed that the running time of MACE directly depends on the efficiency of the SMT solver. As future work we aim to make the proposed method more scalable on large models by investigating recent ideas that have been developed in the context of formal verification of deep neural networks~\cite{cav/HuangKWW17,cav/KatzBDJK17,pacmpl/SinghGPV19} and optimization modulo theories~\cite{NieuwenhuisO06,SebastianiT12}.

\bibliography{aistats2020}
\bibliographystyle{apalike}

\clearpage
\appendix
\section{Background on programming language and program verification} \label{app:model2logic}

\paragraph{Programs}
We assume given a set of function symbols with their arity. For simplicity, we consider the case where operators are untyped and have arity 0 (constants), 1 (unary functions), and 2 (binary functions). We let $c$, $c_1$, and $c_2$ range over constants, unary functions and binary functions respectively. Expressions are built from function
symbols and variables. The set of expressions is defined inductively by the following grammar:
$$\begin{array}{rcll}
e & ::= & x & \mbox{variable} \\
  & \mid & c & \mbox{constant} \\
  & \mid & c_1(e)& \mbox{unary function} \\
  & \mid & c_2(e_1,e_2) & \mbox{binary function}
\end{array}$$
We next assume given a set of atomic predicates. For simplicity, we also consider that predicates have arity 1 or 2, and let $P_1$ and $P_2$ range over unary and binary predicates respectively. We define guards using the following grammar:
$$\begin{array}{rcll}
b & ::= & P_1(e)& \mbox{unary predicate} \\
  & \mid & P_2(e_1,e_2)& \mbox{binary predicate} \\
  & \mid & b_1 \& b_2 & \mbox{conjunction} \\
  & \mid & b_1 \mid\!\mid b_2 & \mbox{disjunction} \\
  & \mid & \neg b & \mbox{negation}
\end{array}$$
We next define commands. These include assignments, conditionals, bounded
loops and return expressions. The set of commands is defined inductively by the following grammar:
$$\begin{array}{rcll}
c & ::= & \mathsf{skip} & \mbox{no-op} \\
  & \mid & x:= e & \mbox{assignment} \\
  & \mid & c_1; c_2 & \mbox{sequential composition} \\
  & \mid & \mathsf{if}~b~\mathsf{then}~c_1~\mathsf{else}~c_2 & \mbox{conditionals} \\
    & \mid & \mathsf{for}~(i=1,\ldots, n)~ \mathsf{do}~c & \mbox{for loop} \\
    & \mid & \mathsf{return}~e & \mbox{return statement}
\end{array}$$
We assume that programs satisfy a well-formedness condition. The condition requires that $\mathsf{return}$ expressions have no successor instruction, i.e.\, we do not allow commands of the form $\mathsf{return}~e;~c$ or $\mathsf{if}~b~\mathsf{then}~c; ~\mathsf{return}~e~\mathsf{else}~c'; c''$. This is without loss of generally, since commands can always be transformed into functionally equivalent programs which satisfy the well-formedness condition.

\paragraph*{Single assignment form}
Our first step to construct characteristic formulae is to transform programs in an intermediate form that is closer to logic. Without loss of generality, we consider loop-free commands, since loops can be fully unrolled. The intermediate form is called a variant of the well-known SSA form~\cite{rosen1988global,cytron1991efficiently} from compiler optimization. Concretely, we transform programs into some weak form of single assignment. This form requires that every non-input variable is defined before being used, and assigned at most once during execution for any fixed input. The main difference with SSA form is that we do not use so-called $\phi$-nodes, as we require that variables are assigned at most once for any fixed input. More technically, our transformation can be seen as a composition of SSA transform with a naive de-SSA transform where $\phi$-nodes are transformed into assignments in the branches of the conditionals.

\paragraph{Path formulae and characteristic formulae}
Our second step is to define the set of path formulae. Informally, a path formula represents a possible execution of the program. Fix a distinguished variable $y$ for return values. Then the path formulae of a command $c$ is defined inductively by the clauses:
\[\arraycolsep=1pt
\begin{array}{rcl}
   \mathrm{PF}_{z:=e}(y) & = & \{z=e\} \\
   \mathrm{PF}_{c_1;c_2}(y) & = & \{ \phi_1 \land \phi_2 \mid
   \phi_1 \in \mathrm{PF}_{c_1}(y) \wedge \\
   & & \quad\quad\quad\quad\quad \phi_2 \in \mathrm{PF}_{c_2}(y)\}  \\
   \mathrm{PF}_{\mathrm{if}~b~\mathrm{then}~c_1~\mathrm{else}~c_2}(y) & = &
   \{ b\wedge \phi_1 \mid \phi_1 \in \mathrm{PF}_{c_1}(y) \} ~ \cup \\
   & & \{ \neg b \wedge \phi_2 \mid \phi_2 \in \mathrm{PF}_{c_2}(y) \} \\
   \mathrm{PF}_{\mathsf{return}~e}(y) & = & \{ y=e \}
    \end{array}
\]
The characteristic formula $\phi_{c}$ of a command $c$ is then defined as:
$$\bigvee_{\phi \in \mathrm{PF}_{c}(y)} \phi$$
One can prove that for every inputs $x_1,\ldots,x_n$, the formula
$\phi_y(x_1,\ldots,x_n,v)$ is valid iff the execution of $c$ on inputs
$x_1,\ldots,x_n$ returns $v$. Note that, strictly speaking, the formula $\phi_y$ contains as free variables the distinguished variable $y$, the inputs $x_1,\ldots,x_n$ of the program, \emph{and} all the program variables, say $z_1\ldots z_m$. However, the latter are fully defined by the characteristic formula so validity of $\phi_y(x_1,\ldots,x_n,v)$
is equivalent to validity of $\exists z_1\ldots z_m.~\phi_y(x_1,\ldots,x_n,v)$.

\newpage

\section{Experiment Details}
\label{app:experiments}
In this section we provide further details on the detasets and methods used in or experiments, together with some additional results.

\subsection{Model Selection}
\label{app:model_selection}
To demonstrate the flexibility of our approach, we explored four different differentiable and non-differentiable model classes, i.e., decision tree, random forest, logistic regression and  multilayer perceptron (MLP). As the main focus of our work is to generate counterfactuals for a broad range of already trained models, we  opted for  models' parametrization that result in good performance on the considered datasets (e.g., default parameters).
For instance, for the MLP, we opted for two hidden layers with 10 neurons, since it present better performance in the Adult dataset ($\%82.52/\%81.94$ training/test accuracy) than other architectures with $\textrm{hidden}=\{100\} \textrm{(default})$ and  $\textrm{hidden}=\{100,100\}$ which result in $\%81.69/\%81.06$ and  $\%81.51/\%80.82$ training/test accuracy, respectively.
We leave the exploration of other datasets (larger feature spaces), more complex models (deeper MLPs) and other SMT solvers as future work.

\subsection{Datasets}
\label{app:datasets}
Here we detail the different types of variables present in each dataset. We used the default features for the Adult and COMPAS datasets, and applied the same preprocessing used in \cite{ustun2019actionable} for the Credit dataset. All samples with missing data were dropped. We remark that we have relied on broadly studied datasets in the literature on fairness and interpretability of ML for consequential decision making. For instance, the  Credit dataset [34] ($n=29,623, d=14$) has been previously studied by the Actionable Recourse work [29], and the Adult [1] ($n=45,222, d=12, d \textrm{(one-hot)}=51$) and COMPAS [18] ($n=5,278, d=5, d \textrm{(one-hot)}=7$) have been previously used in the context of fairness in ML [Joseph et al., 2016; Zafar et al., 2017; Agarwal et al. 2018].

\textbf{Adult} ($n=45,222, d=12, d \textrm{(one-hot)}=51$):
\begin{itemize}
  \item Integer: Age, Education Number, Hours Per Week
  \item Real: Capital Gain, Capital Loss
  \item Categorical: Sex, Native Country, Work Class, Marital Status, Occupation, Relationship
  \item Ordinal: Education Level
\end{itemize}

\textbf{Credit} ($n=29,623, d=14, d \textrm{(one-hot)}=20$):
\begin{itemize}
  \item Integer: Total Overdue Counts, Total Months Overdue, Months With Zero Balance Over Last 6 Months, Months With Low Spending Over Last 6 Months, Months With High Spending Over Last 6 Months
  \item Real: Max Bill Amount Over Last 6 Months, Max Payment Amount Over Last 6 Months, Most Recent Bill Amount, Most Recent Payment Amount
  \item Categorical: Is Male, Is Married, Has History Of Overdue Payments
  \item Ordinal: Age Group, Education Level
\end{itemize}

\textbf{COMPAS} ($n=5,278, d=5, d \textrm{(one-hot)}=7$):
\begin{itemize}
  \item Integer: -
  \item Real: Priors Count
  \item Categorical: Race, Sex, Charge Degreee
  \item Ordinal: Age Group
\end{itemize}

\subsection{Handling Mixed Data Types}
\label{app:handling_mixed_data_types}
While the proposed approach (MACE) naturally handles mixed data types, other approaches do not. Specifically, the Feature Tweaking method generates counterfactual explanations for Random Forest models trained on non-hot embeddings of the dataset, meaning that the resulting counterfactuals will not have multiple categories of the same variable activated at the same time. However, because this method is only restricted to working with real-valued variables, the resulting counterfactual is must undergo a post-processing step to ensure integer-, categorical-, and ordinal-based variables are plausible in the counterfactual. The Actionable Recourse method, on the other hand, explanations for Logistic Regression models trained on one-hot embeddings of the dataset, hence requiring additional constraints to ensure that multiple categories of a categorical variable are not simultaneously activated in the counterfactual. While the authors suggest how this can be supported using their method, their open-source implementation \textit{converts categorical columns to binary where possible and drops other more complicated categorical columns}, postponing to future work. Furthermore, the authors state that \textit{the question of mutually exclusive features will be revisited in later releases}~\footnote{\url{https://github.com/ustunb/actionable-recourse/blob/master/examples/ex_01_quickstart.ipynb}}. Moreover, ordinal variables are not supported using this method. The overcome these shortcomings, the counterfactuals generated by both approaches is post-processed to ensure correctness of variable types by rounding integer-based variables, and taking the maximally activated category as the counterfactual category.

\newpage
\section{Additional Results}
\label{app:results}

\begin{table*}[h]
\scriptsize
  \setlength{\tabcolsep}{2pt}
  \caption{Wall-clock time (seconds) for computing the nearest counterfactual explanation (without constraints). $N = \Omega_\text{MACE} \cap \Omega_\text{Other}$ factual samples; cells are shaded for unsupported tests. Lower run-time is better. The run-time for MACE depends on $O(\log(1/\epsilon))$, i.e., orders of magnitude more accuracy only cost linearly more run-time. These results should be considered along Tables \ref{table:CoverageComparison}, \ref{table:DistanceComparison} comparing coverage $\Omega$ and distance $\delta$.}
  \label{table:RunTimeUnconstrained}
  \begin{adjustwidth}{-.4in}{-.4in} 
  \begin{tabular}{|l|l||c|c|c|c|c|c|c|c|c|}
    \cline{3-11}
    \multicolumn{2}{l|}{\multirow{2}{*}{}}                 & \multicolumn{3}{|c|}{Adult}                                              & \multicolumn{3}{|c|}{Credit}                                               & \multicolumn{3}{|c|}{COMPAS}                                   \\ \cline{3-11}
    \multicolumn{2}{l|}{}                                  & $\ell_0$               & $\ell_1$               & $\ell_\infty$          & $\ell_0$               & $\ell_1$               & $\ell_\infty$            & $\ell_0$           & $\ell_1$             & $\ell_\infty$      \\ \hline \hline
    \multirow{5}{*}{tree}    & MACE ($\epsilon = 10^{-1}$) & \zpl{5.65}{2.18}       & \zpl{3.01}{0.74}       & \zpl{3.47}{0.93}       & \zpl{3.48}{1.25}       & \zpl{3.44}{1.70}       & \zpl{2.39}{0.64}         & \zpl{2.41}{1.06}   & \zpl{1.22}{0.36}     & \zpl{1.62}{0.78}   \\ \cline{3-11}
                             & MACE ($\epsilon = 10^{-3}$) & \zpl{17.59}{4.87}      & \zpl{9.58}{3.05}       & \zpl{10.43}{2.98}      & \zpl{15.84}{4.78}      & \zpl{7.55}{3.44}       & \zpl{4.44}{2.20}         & \zpl{7.07}{2.09}   & \zpl{5.72}{1.28}     & \zpl{4.99}{1.89}   \\ \cline{3-11}
                             & MACE ($\epsilon = 10^{-5}$) & \zpl{35.32}{14.07}     & \zpl{20.35}{6.34}      & \zpl{20.44}{9.55}      & \zpl{25.47}{8.71}      & \zpl{18.46}{6.24}      & \zpl{10.58}{6.36}        & \zpl{13.49}{6.44}  & \zpl{9.22}{4.21}     & \zpl{10.76}{4.60}  \\ \cline{3-11}
                             & MO                          & \zpl{1.04}{0.26}       & \zpl{0.85}{0.27}       & \zpl{0.87}{0.22}       & \zpl{0.53}{0.15}       & \zpl{0.64}{0.26}       & \zpl{0.54}{0.23}         & \zpl{0.15}{0.07}   & \zpl{0.12}{0.06}     & \zpl{0.16}{0.07}   \\ \cline{3-11}
                             & PFT                         & \zshade                & \zshade                & \zshade                & \zpl{1.45}{0.42}       & \zpl{1.50}{0.36}       & \zpl{1.91}{0.79}         & \zpl{0.12}{0.05}   & \zpl{0.13}{0.06}     & \zpl{0.12}{0.05}   \\ \hline
    \multirow{5}{*}{forest}  & MACE ($\epsilon = 10^{-1}$) & \zpl{27.98}{9.48}      & \zpl{17.68}{4.82}      & \zpl{19.05}{6.11}      & \zpl{28.12}{9.31}      & \zpl{21.88}{10.04}     & \zpl{21.47}{11.07}       & \zpl{8.07}{3.36}   & \zpl{3.18}{1.15}     & \zpl{3.52}{1.93}   \\ \cline{3-11}
                             & MACE ($\epsilon = 10^{-3}$) & \zpl{69.19}{15.76}     & \zpl{55.79}{15.78}     & \zpl{52.31}{15.39}     & \zpl{57.29}{26.69}     & \zpl{40.75}{17.85}     & \zpl{26.21}{11.71}       & \zpl{15.05}{5.15}  & \zpl{10.75}{3.03}    & \zpl{8.53}{3.55}   \\ \cline{3-11}
                             & MACE ($\epsilon = 10^{-5}$) & \zpl{89.81}{28.99}     & \zpl{84.89}{35.14}     & \zpl{78.49}{23.85}     & \zpl{107.83}{52.32}    & \zpl{90.04}{38.02}     & \zpl{72.38}{37.77}       & \zpl{33.26}{9.79}  & \zpl{19.95}{10.03}   & \zpl{17.22}{7.90}  \\ \cline{3-11}
                             & MO                          & \zpl{1.14}{0.35}       & \zpl{0.98}{0.25}       & \zpl{0.94}{0.36}       & \zpl{0.80}{0.27}       & \zpl{0.80}{0.35}       & \zpl{0.80}{0.28}         & \zpl{0.16}{0.06}   & \zpl{0.17}{0.08}     & \zpl{0.15}{0.07}   \\ \cline{3-11}
                             & PFT                         & \zshade                & \zshade                & \zshade                & \zpl{13.41}{7.09}      & \zpl{10.46}{4.67}      & \zpl{11.79}{6.51}        & \zpl{1.93}{0.81}   & \zpl{2.11}{1.07}     & \zpl{1.83}{0.87}   \\ \hline
    \multirow{5}{*}{lr}      & MACE ($\epsilon = 10^{-1}$) & \zpl{0.85}{0.29}       & \zpl{0.66}{0.26}       & \zpl{0.74}{0.29}       & \zpl{0.33}{0.15}       & \zpl{1.17}{1.79}       & \zpl{0.49}{0.30}         & \zpl{0.21}{0.10}   & \zpl{0.19}{0.10}     & \zpl{0.22}{0.11}   \\ \cline{3-11}
                             & MACE ($\epsilon = 10^{-3}$) & \zpl{2.22}{0.86}       & \zpl{3.55}{1.50}       & \zpl{5.15}{3.51}       & \zpl{0.87}{0.20}       & \zpl{10.57}{8.14}      & \zpl{6.11}{3.51}         & \zpl{0.52}{0.18}   & \zpl{0.31}{0.12}     & \zpl{0.54}{0.20}   \\ \cline{3-11}
                             & MACE ($\epsilon = 10^{-5}$) & \zpl{2.73}{0.73}       & \zpl{6.60}{3.01}       & \zpl{13.32}{6.70}      & \zpl{1.19}{0.56}       & \zpl{25.10}{21.67}     & \zpl{16.21}{8.84}        & \zpl{0.84}{0.22}   & \zpl{0.72}{0.28}     & \zpl{0.77}{0.21}   \\ \cline{3-11}
                             & MO                          & \zpl{7.52}{1.91}       & \zpl{6.62}{1.73}       & \zpl{5.73}{1.14}       & \zpl{1.86}{0.82}       & \zpl{1.41}{0.53}       & \zpl{1.69}{0.79}         & \zpl{0.30}{0.22}   & \zpl{0.25}{0.12}     & \zpl{0.25}{0.11}   \\ \cline{3-11}
                             & AR                          & \zshade                & \zpl{2.05}{0.45}       & \zpl{1.86}{0.03}       & \zshade                & \zpl{0.72}{0.15}       & \zpl{0.66}{0.07}         & \zshade            & \zpl{0.07}{0.01}     & \zpl{0.06}{0.01}   \\ \hline
    \multirow{4}{*}{mlp}     & MACE ($\epsilon = 10^{-1}$) & \zpn{2586}{4523}       & \zpn{8070}{5995}       & \zpn{5091}{6616}       & \zpn{1743}{4171}       & \zpn{3432}{5615}       & \zpn{10309}{10088}       & \zpn{59}{53}       & \zpn{158}{135}       & \zpn{90}{90}       \\ \cline{3-11}
                             & MACE ($\epsilon = 10^{-3}$) & \zpn{4187}{9899}       & \zpn{34101}{29853}     & \zpn{7094}{10919}      & \zpn{1703}{5889}       & \zpn{3304}{4944}       & \zpn{8689}{11638}        & \zpn{79}{55}       & \zpn{180}{139}       & \zpn{122}{103}     \\ \cline{3-11}
                             & MACE ($\epsilon = 10^{-5}$) & \zpn{5888}{9760}       & \zpn{44470}{30907}     & \zpn{19712}{14117}     & \zpn{1901}{4892}       & \zpn{4736}{5080}       & \zpn{11129}{9773}        & \zpn{100}{56}      & \zpn{257}{168}       & \zpn{203}{149}     \\ \cline{3-11}
                             & MO                          & \zpl{6.66}{2.17}       & \zpl{6.61}{1.96}       & \zpl{6.40}{1.60}       & \zpl{2.02}{2.09}       & \zpl{2.43}{0.41}       & \zpl{1.90}{0.83}         & \zpl{0.35}{0.12}   & \zpl{0.45}{0.10}     & \zpl{0.32}{0.09}   \\ \hline
  \end{tabular}
  \end{adjustwidth}
\end{table*}

\begin{table*}[t]
  \setlength{\tabcolsep}{3pt}
  \caption{Percentage of factual samples for which the nearest counterfactual sample requires a reduction in age for a random forest trained on the Adult dataset, and the corresponding increase in distance to nearest counterfactual when restricting the approaches not to reduce age: $100 \times \mathbb{E}[\delta_\text{restr.} / \delta_\text{unrestr.} - 1]$. 
  }
  \label{table:AgeReduction}
  \centering
  \small
  \begin{tabular}{|l||c|c|c|c|c|c|}
    \cline{2-7}
    \multicolumn{1}{l|}{} & \multicolumn{2}{|c|}{$\ell_0$}      & \multicolumn{2}{|c|}{$\ell_1$}      & \multicolumn{2}{|c|}{$\ell_\infty$} \\ \cline{2-7}
    \multicolumn{1}{l|}{} & $\%$ age-red. & rel. dist. increase & $\%$ age-red. & rel. dist. increase & $\%$ age-red. & rel. dist. increase \\ \hline\hline
    MACE ($\epsilon = 10^{-5}$)  & \zpm{3.6}   & \zpm{0}               & \zpm{7.4}   & \zpm{61.3}            & \zpm{34.2}  & \zpm{13.9}            \\ \cline{2-7}
    MO                    & \zpm{24.6}  & \zpm{29.7}            & \zpm{34.6}  & \zpm{94.6}            & \zpm{34.2}  & \zpm{66.6}            \\ \hline
  \end{tabular}
\end{table*}

\subsection{Comprehensive Distance $\zvec{\delta}$ Results}
\label{app:runtime_analysis}
\begin{figure*}[h]
  \centering
  \centerline{\includegraphics[width=0.9\textwidth]{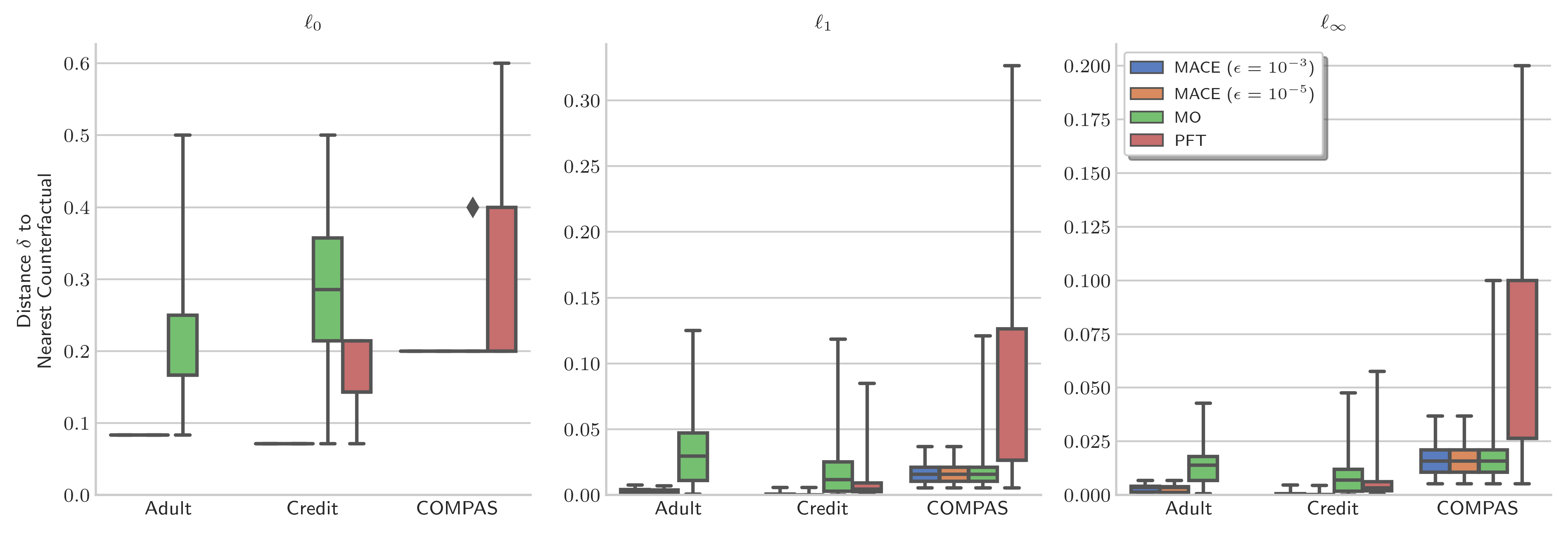}}
  \centerline{\includegraphics[width=0.9\textwidth]{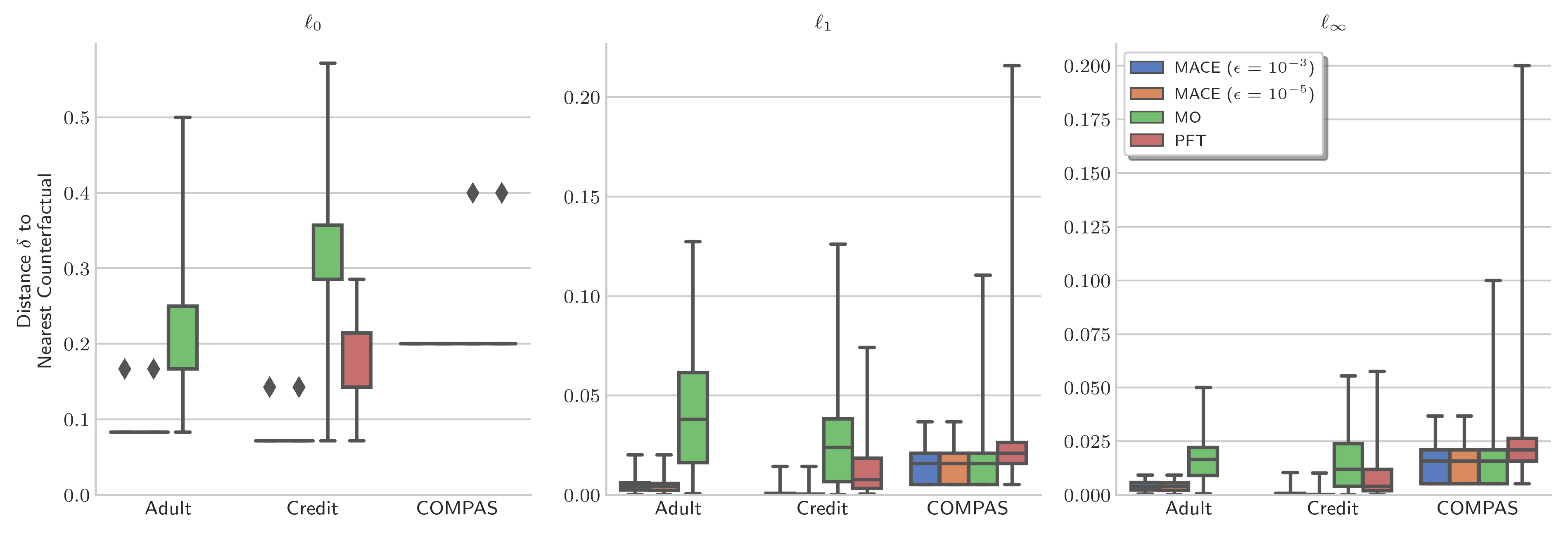}}
  \centerline{\includegraphics[width=0.9\textwidth]{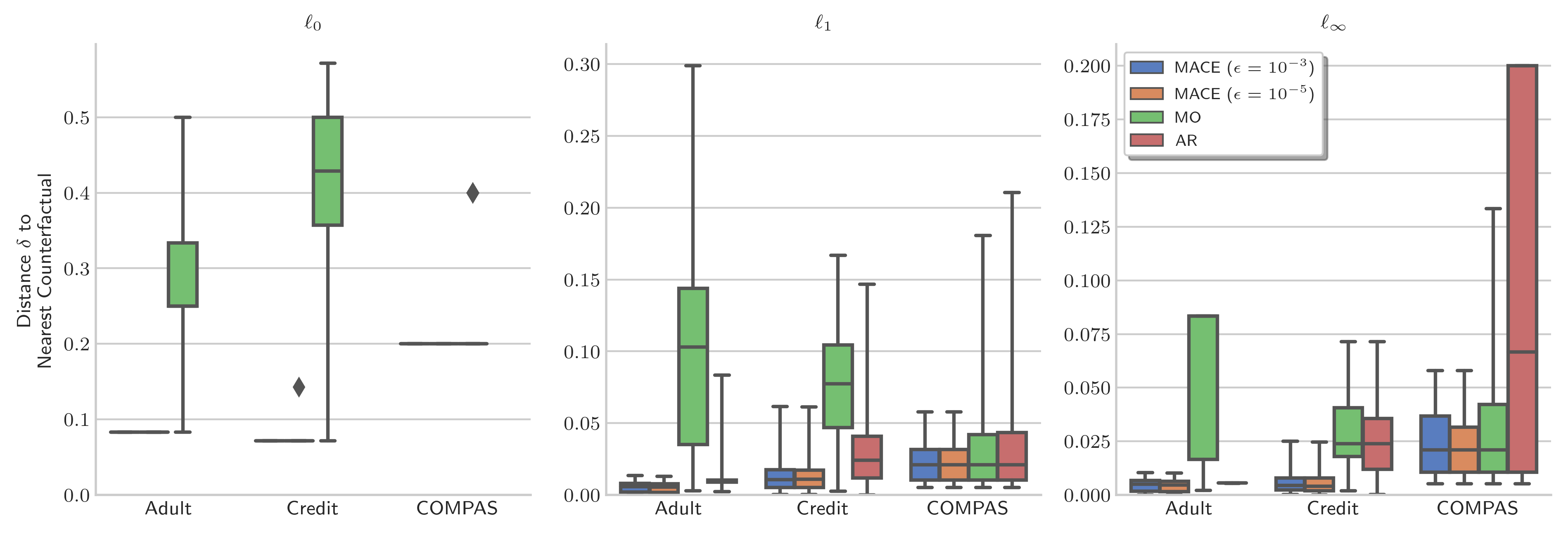}}
  \centerline{\includegraphics[width=0.9\textwidth]{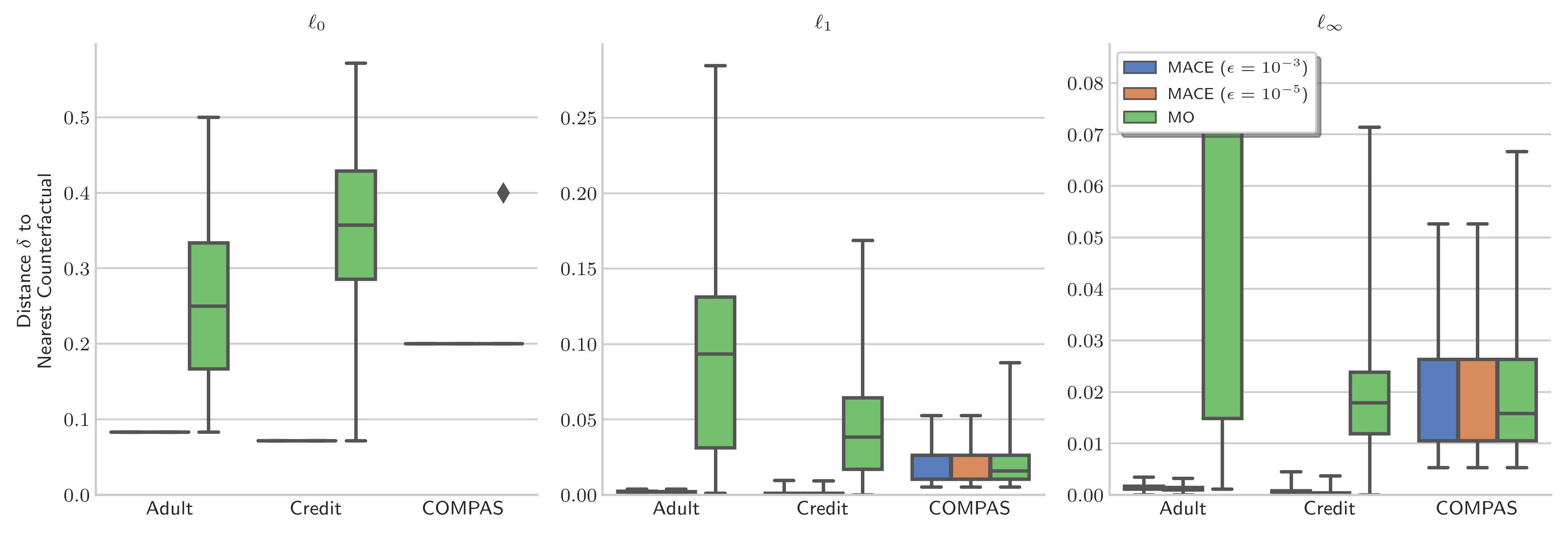}}
  \caption{Comparison of approaches for generating unconstrained counterfactual explanations for a \textbf{(top to bottom) trained decision tree, random forest, logistic regression, and multilayer perceptron model}. Here the distribution of distance $\zvec{\delta}$ is shown \emph{upon termination} of Algorithm \ref{alg:bin-search}; lower distance is better. For each bar, $N = 500 \times \Omega$ from Table \ref{table:CoverageComparison}, and absent bars refer to $\Omega = 0$. In all setups, MACE results are at least as good as any other approach.}
  \label{figure:DistanceComparison}
\end{figure*}

Following the presentation of coverage $\Omega$ results in Table \ref{table:CoverageComparison} and relative distance $\zvec{\delta}$ improvement (reduction) in Table \ref{table:DistanceComparison} of the main body, in Figure \ref{figure:DistanceComparison} we present the complete distribution of counterfactual distances upon termination of Algorithm \ref{alg:bin-search}. Importantly, we see that in all setups (approaches $\times$ models $\times$ norms $\times$ datasets), MACE results are at least as good as any other approach (MO, PFT, AR).

\subsection{Quality vs Complexity}
\label{app:quality_vs_complexity}
\begin{figure*}[h]
  \centering
  \centerline{\includegraphics[width=\textwidth]{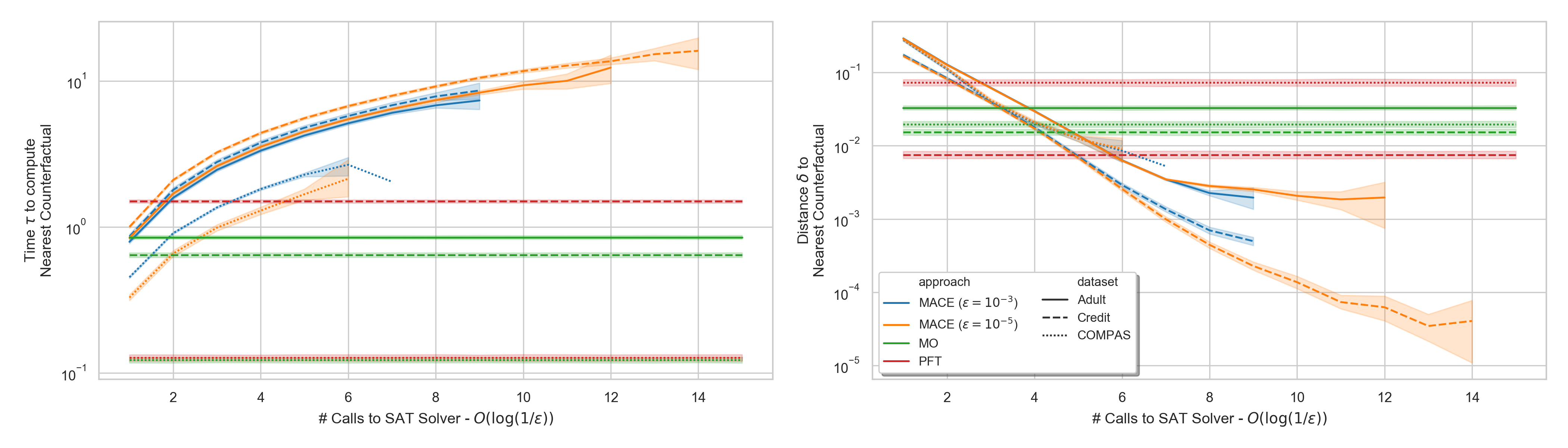}}
  \centerline{\includegraphics[width=\textwidth]{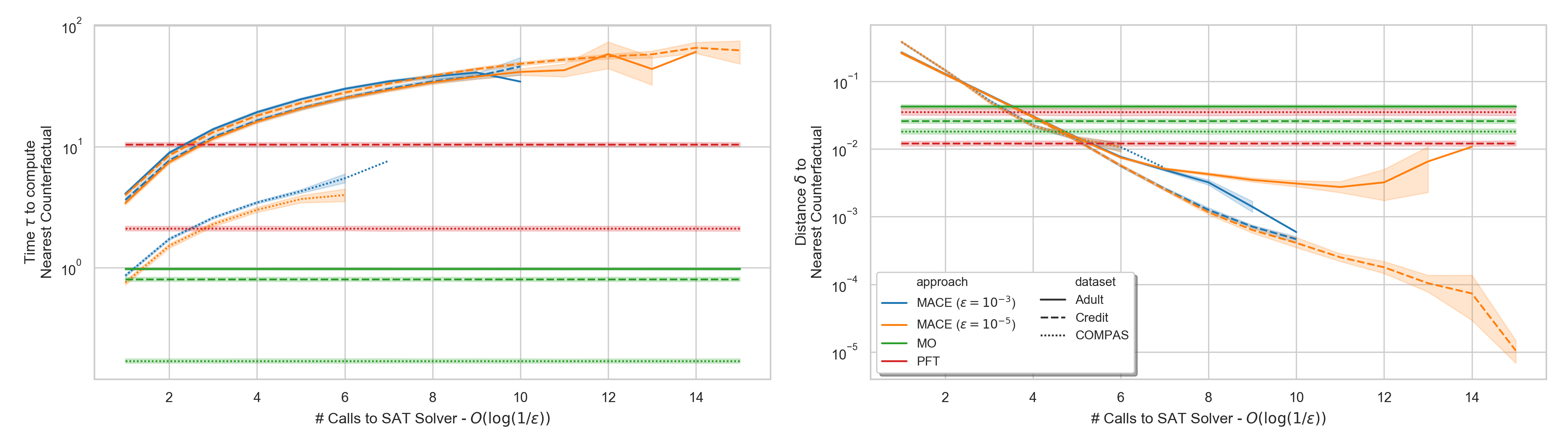}}
  \centerline{\includegraphics[width=\textwidth]{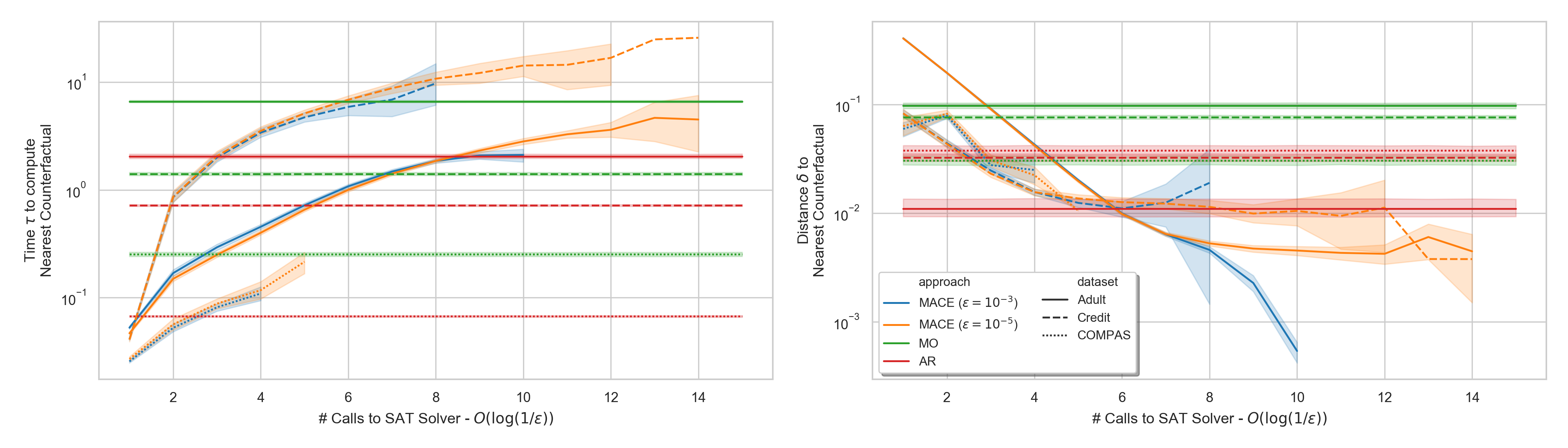}}
  \centerline{\includegraphics[width=\textwidth]{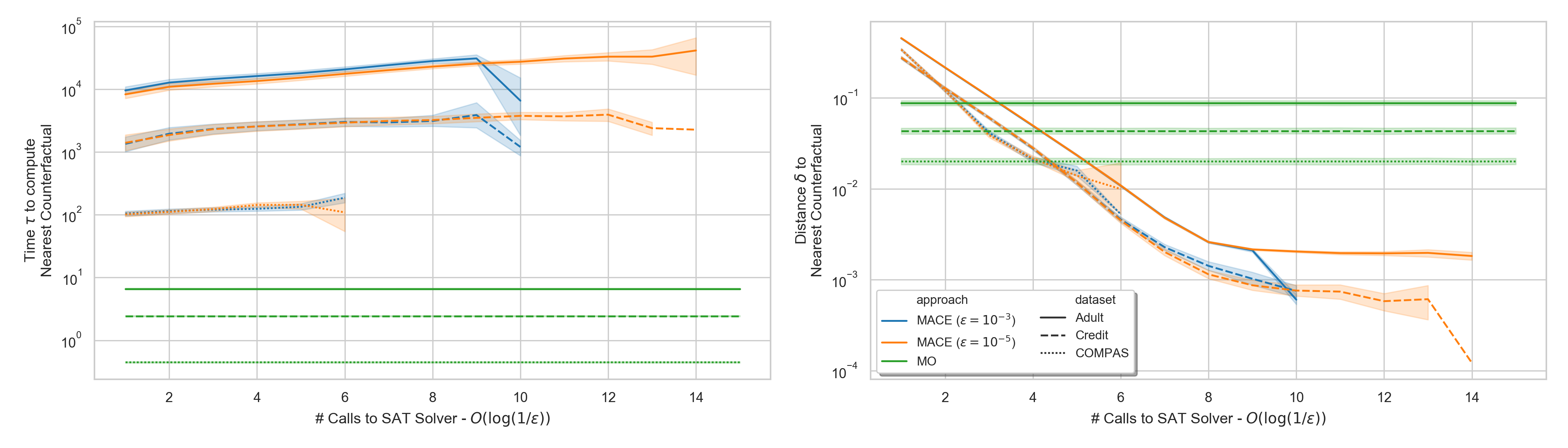}}
  \caption{Comparison of approaches for generating unconstrained counterfactual explanations for a \textbf{(top to bottom) trained decision tree, random forest, logistic regression, and multilayer perceptron model}. Here the average distance $\zvec{\delta}$ and run-time $\tau$ is shown \emph{during execution} of Algorithm \ref{alg:bin-search} (i.e., over number of calls to the $\SAT$ solver); lower distance and lower run-time is better. Other approaches (MO, PFT, AR) would only be shown as a single point on these plots, and therefore we repeat their results over all values of the x-axis for ease of comparison against MACE. Results are averaged over all plausible counterfactuals ($N = 500 \times \Omega$ from Table \ref{table:CoverageComparison},). As expected, Algorithm \ref{alg:bin-search} terminates after different number of iterations depending on the factual instance; this explains the observed larger variance in results for higher number of iterations. These results confirm our intuition: not only does MACE always achieve a lower counterfactual distance upon termination, in many cases an early termination of MACE generates closer counterfactuals while also being less computationally demanding.}
  \label{figure:tradeoff_unconstrained}
\end{figure*}

In the main text and in the previous section, we considered distance comparisons \emph{upon termination} of Algorithm \ref{alg:bin-search}; in this section we explore the effect of the accuracy parameter $\epsilon$ jointly on quality (distance $\zvec{\delta}$) and complexity (run-time $\tau$) \emph{during execution} of Algorithm \ref{alg:bin-search}. Importantly, the number of calls made to the $\SAT$ solver follows $O(\log(1/\epsilon))$, where $\epsilon$ is the desired the accuracy term, i.e., orders of magnitude more accuracy only cost linearly more $\SAT$ calls. 
The run-time of each call to the $\SAT$ solver is governed by a number of parameters, including the implementation details of the $\SAT$ solver\footnote{This is assumed beyond the scope of the paper; we built MACE atop the open-source PySMT library \cite{pysmt2015} with the Z3 \cite{DBLP:conf/tacas/MouraB08} backend to demonstrate its model-agnostic support of off-the-shelf models.}, the compute hardware\footnote{All tests were conducted using one X86\_64 Xeon(R) CPU @ 2.60GHz, and 8GB memory.}, among other factors.
Clearly, a higher desired accuracy (i.e., $\epsilon \rightarrow 0$) will result in closer counterfactuals ($\zvec{\delta} \in [\zvec{\delta}^*, \zvec{\delta}^* + \epsilon]$) at the cost of higher run-time (higher $\tau$), while leaving the coverage $\Omega$ unchanged (remaining at $100\%$, by design).
Figure \ref{figure:tradeoff_unconstrained} depicts the average counterfactual distance and average run-time against the number of calls to the $\SAT$ solver, confirming the intuition above: not only does MACE always achieve a lower counterfactual distance\footnote{Reminder: lower distance is more desirable, as it specifies the least change required of the individual's features.} upon termination, in many cases an early termination of MACE generates closer counterfactuals while also being less computationally demanding.

In addition to studying the quality vs complexity tradeoff against number of calls to the $\SAT$ solver, in Table \ref{table:RunTimeUnconstrained} we compare final run-times (in seconds) upon-termination of Algorithm \ref{alg:bin-search} for various setups. The results show that MACE takes less than 5 seconds for logistic regression; between 5 and 60 seconds for decision trees and random forests; and between one minute and three hours for the multilayer perceptron (outliers were not excluded in computed mean runtimes).
In contrast, competing approaches (MO, PFT, AR) require at most 30 seconds to generate a counterfactual explanation, when possible (note that the coverage for AR and PFT is often significantly below 100\%, and only MACE is able to \emph{generate} counterfactuals for the multilayer perceptron; MO requires access to the training data as it searches through the training set for a counterfactual).
We believe that this difference is compensated (at least for the decision tree, the random forest, and the logistic regression classifiers) by the main properties of MACE compared to previous works, i.e.: i) \emph{model-agnostic} (\{non-\}linear, \{non-\}differentiable, \{non-\}convex); ii) \emph{data-agnostic} (heterogeneous features); iii) \emph{provable closeness guarantees}; and iv) \emph{100\% coverage}, even under plausibility and diversity constraints.
Regarding the results on MLPs, we are well aware of prior work that develops efficient SMT-based methods for verifying large deep neural networks (see formal verification of deep neural networks~\cite{cav/HuangKWW17,cav/KatzBDJK17,pacmpl/SinghGPV19} and optimization modulo theories~\cite{NieuwenhuisO06,SebastianiT12}); indeed we plan to leverage state-of-the-art tools to improve the efficiency of our implementation, in particular for MLP-based models. With the current implementation of MACE, our main goal was to explore the use of off-the-shelf SMT-solvers already available in Python to generate counterfactuals in a broad range of settings, justifying our lesser emphasis on efficiency.

In practice the choice of epsilon should reflect the desired distance granularity from the operator, the number and range of attributes in the data space, and the decided upon distance norm. For example, using the $\ell_0$ norm, which tracks the number of attributes changed, the lowest achievable distance granularity is $1/J$ where $J$ is the data dimensionality. Therefore, choosing any $\epsilon < 1 / J$ is sufficient and will result in the optimal counterfactual for this choice of distance metric. As another example, for the continuous $\ell_1$ norm, too much granularity may result in a lack of trust for the end-user -- consider the adult dataset with account balance feature with range $R = \$50,000$; choosing a fine granularity may result in a counterfactual that suggests that only a few dollars change in the account balance can flip the prediction (e.g., result in the approval of a loan). It is important to point out that this phenomenon is not a fault of the counterfactual generating method (i.e., MACE), but of the robustness of the underlying classifier and its decision boundary. While such an explanation may not be favorable for an end-user, it may assist a system administrator or model designer to assay the robustness and safety of their model prior to deployement.

\subsection{Additional Constrained Results}
\label{app:additional_constrained_results}

Following the study of counterfactuals that change or reduce age (Section \ref{sec:experiments}), we regenerate counterfactual explanations for those samples for which age-reduction was required, with an additional plausibility constraint ensuring that the age shall not decrease. The results presented in Table \ref{table:AgeReduction} show interesting results. Once again, we observe that the additional plausibility constraint for the age incurs significant increases in the distance of the nearest counterfactual -- being, as expected, more pronounced for the $\ell_1$ and the $\ell_\infty$ norms. For the $\ell_0$ norm, we find that for the 18 factual samples (i.e., $3.6\% \times 500$) for which the unrestricted MACE required age-reduction, the addition of the no-age-reduction constraint results in counterfactuals at the same distance, while suggesting a change in work class ($5/18$) or education level ($4/18$) instead of changing age.

\subsection{Details on diverse counterfactuals example}
\label{app:diverse_counterfactuals_details}

In the main body, we described a scenario where a logistic regression model had predicted that a loan borrower, John, would default on his loan. Here is john's complete feature list: John is a married male between 40-59 years of age with some university degree. Over the last 6 months, Max Bill Amount = 500.0, Max Payment Amount = 60.0, Months With Zero Balance = 0.0, Months With Low Spending = 0.0, Months With High Spending = 1.0. Furthermore, John has a history of overdue payments, his Most Recent Bill Amount = 370.0, and his Most Recent Payment Amount = 40.0

\end{document}